\pdfoutput=1
\documentclass[11pt]{article}

\usepackage[final]{acl}

\usepackage{times}
\usepackage{latexsym}
\usepackage{kotex}
\usepackage{multicol}
\usepackage{booktabs}
\usepackage{tabularx}
\usepackage{multirow, makecell}
\usepackage{colortbl}
\usepackage{arydshln}
\usepackage{array}
\usepackage{graphicx}
\usepackage{amssymb} 
\definecolor{headercolor}{HTML}{E0EBF6}% Light blue
\definecolor{conceptcolor}{HTML}{E5F0DB} % Light green
\definecolor{answerbackcolor}{HTML}{F7E6D8} % Light orange

\usepackage[T1]{fontenc}
\usepackage[utf8]{inputenc}

\usepackage{microtype}

\usepackage{inconsolata}

\usepackage{graphicx}
\newcommand\blfootnote[1]{%
  \begingroup
  \renewcommand\thefootnote{}\footnote{#1}%
  \addtocounter{footnote}{-1}%
  \endgroup
}
\title{The Impact of Negated Text on Hallucination with Large Language Models}

\author{
  Jaehyung Seo$^{1}$, Hyeonseok Moon$^{1}$ and Heuiseok Lim$^{1\dagger}$ \\
  $^{1}$Department of Computer Science and Engineering, Korea University \\
  \texttt{\{seojae777,glee889,limhseok\}@korea.ac.kr}
}

\begin{document}
\maketitle
\begin{abstract}
\blfootnote{$^\dagger$ Corresponding Author}
Recent studies on hallucination in large language models (LLMs) have been actively progressing in natural language processing. However, the impact of negated text on hallucination with LLMs remains largely unexplored. In this paper, we set three important yet unanswered research questions and aim to address them. To derive the answers, we investigate whether LLMs can recognize contextual shifts caused by negation and still reliably distinguish hallucinations comparable to affirmative cases. We also design the NegHalu dataset by reconstructing existing hallucination detection datasets with negated expressions. Our experiments demonstrate that LLMs struggle to detect hallucinations in negated text effectively, often producing logically inconsistent or unfaithful judgments. Moreover, we trace the internal state of LLMs as they process negated inputs at the token level and reveal the challenges of mitigating their unintended effects.

%We suggest an editing approach leveraging knowledge written in negated forms with the same meaning for mitigating hallucinations in negated contexts and highlighting the importance of refining model architectures to better understand and process negation. In this paper, we investigate whether LLMs can reliably maintain their hallucination detection capabilities when presented with negated text as context. 

\end{abstract}

\section{Introduction}
The rapid advancement of large language models (LLMs) continues to drive the release of diverse open-source models capable of performing a wide range of tasks \citep{touvron2023llama,jiang2023mistral,team2024gemma}. As these models become more prevalent, the ability to distinguish whether generated outputs contain hallucinations is becoming increasingly critical \citep{magesh2024hallucination}. Detecting hallucination involves identifying content that is either contextually unfaithful or contradictory to real-world facts and assessing the truthfulness of such outputs \citep{ji2023survey,zhang2023siren,huang2023survey}.

\begin{figure}[t]
\centering
\includegraphics[width=1\linewidth]{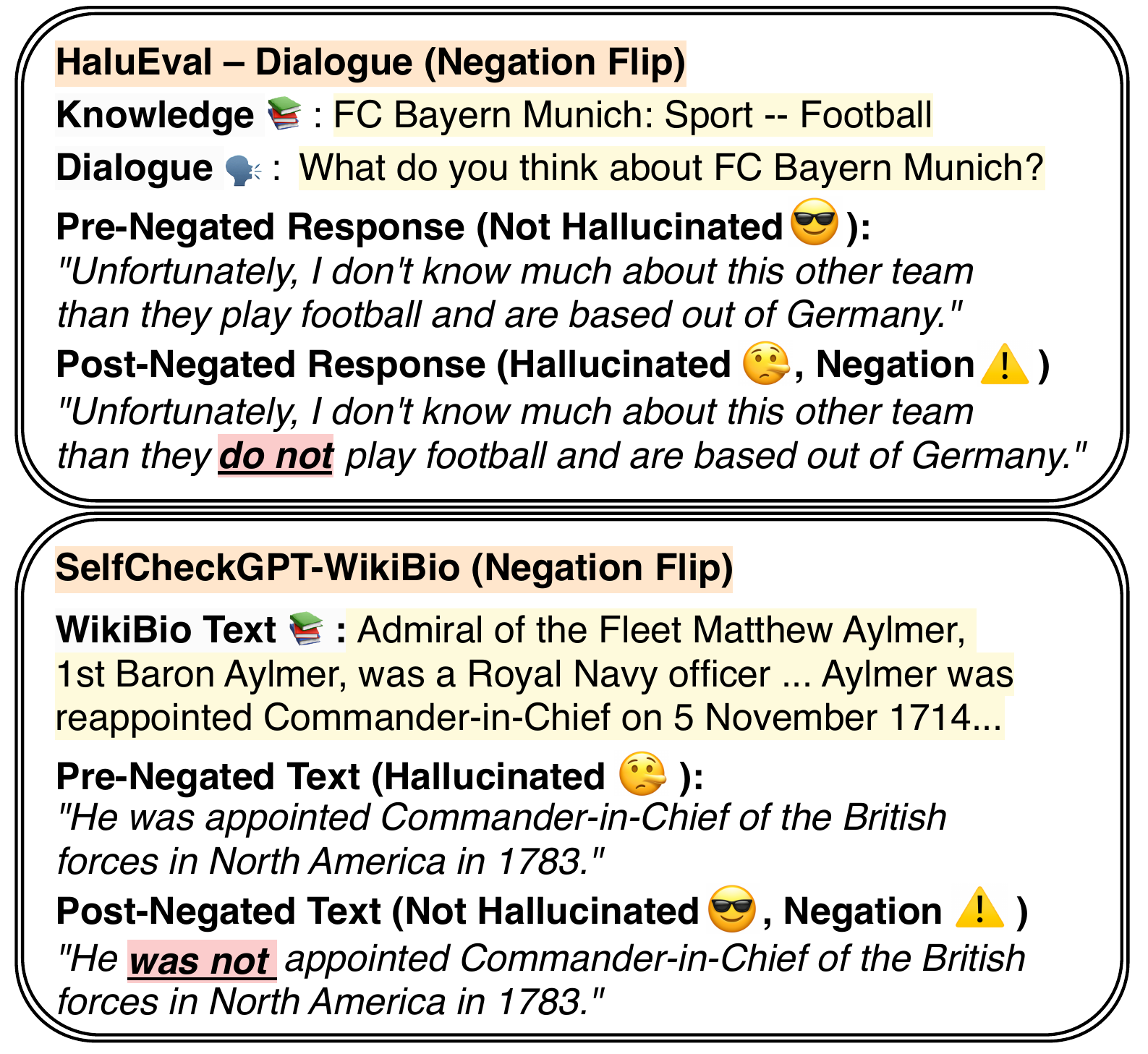}
\caption{Examples illustrating how negation can flip the hallucination label. The negated response introduces or resolves contradictions with the given knowledge.}
 \label{fig:intro_figure}
\end{figure}

% (1) 거대언어모델이 급격하게 발전하면서, 오픈 소스로 다양한 모델들이 공개되고 다양한 태스크를 수행하기 시작했다 \citep{touvron2023llama,jiang2023mistral,team2024gemma}.  As these models become more prevalent, the need for 생성된 결과물에 환각이 포함되어 있는지 구분하는 능력과 태스크 grows increasingly critical \citep{magesh2024hallucination}. \citep{magesh2024hallucination}. Hallucination detection은 맥락적으로 일관성이 없거나 실제 세계의 사실과 상충하는 내용을 식별하고, 이를 기반으로 해당 진술의 진실성을 판단하는 과정을 의미한다 \citep{ji2023survey,zhang2023siren,huang2023survey}. 

Recent research on hallucination detection actively focuses on improving the reliability of LLMs by identifying their limitations and refining models based on insights into hallucinated outputs \citep{manakul2023selfcheckgpt,jiang2024large,cheninside}. However, there is a limited exploration of how negated text affects hallucination detection with LLMs. Negated text, which includes negation markers (e.g., "not," "never," "no," "without"), is commonly used in everyday communication \citep{gubelmann2022context,hossain2022analysis}. Although these markers are typically single tokens, they exert a disproportionately large influence on the overall processing factuality of a sentence, fundamentally altering its meaning \citep{vanek2024mental}. 
However, unlike humans, LLMs struggle to effectively handle negation and infer contextual meaning \citep{truong2023language,ye2023assessing}. Moreover, negative knowledge can introduce hallucinations into commonsense reasoning \citep{chen2023say,seo2024kocommongen}, as LLMs tend to misrepresent negation as a faulty logical operator, leading to severe hallucinations \citep{bhar2024strong}.
% (2) 최근 hallucination detection 연구들은 LLM의 신뢰성을 확보하고 한계를 파악하여 이를 기반으로 모델을 개선하며, 환각 정보를 판별하기 위해 활발하게 진행되고 있었다 \citep{manakul2023selfcheckgpt,jiang2024large,cheninside}. 그러나, 부정형 텍스트가 입력으로 주어지는 경우 LLM의 hallucination detection에 어떠한 영향을 줄 수 있는지에 대한 연구는 부족하다. 부정형 텍스트는 부정어 (e.g.,)를 포함하며 실제 세계의 사람들의 발화에서 빈번하게 사용되고 있다 \citep{gubelmann2022context,hossain2022analysis}.  Although these markers are typically single tokens, they exert a disproportionately large influence on the overall processing factuality of a sentence, fundamentally altering its meaning \citep{vanek2024mental}.그러나 언어 모델들은 사람과 다르게 negation을 효과적으로 handle하고 맥락을 추론하는데 어려움을 겪고 있다 \citep{truong2023language,ye2023assessing}. 더욱이 부정형으로 표현된 knowledge는 언어 모델의 상식 추론에 hallucination을 생성하고 있으며 \citep{chen2023say,seo2024kocommongen}, LLMs는 부정어를 faulty meaning representations of logical operators하여 강력한 환각을 야기하고 있다\citep{bhar2024strong}.   

In this paper, we present an initial exploration of how negated text influences hallucination detection in LLMs. We propose three open research questions and work towards answering them.

\begin{itemize}

\item \textbf{RQ1.} Can LLMs distinguish between hallucinations and faithful statements in negated text as effectively as in affirmative text?

\item \textbf{RQ2.} Can the model internally recognize differences caused by negation when detecting hallucinations?

\item \textbf{RQ3.} Can targeted intervention strategies improve hallucination detection in the negated text?

\end{itemize}

To address these research questions, we investigate whether LLMs can recognize contextual shifts and reliably detect hallucinations in the presence of negated text. As illustrated in Figure~\ref{fig:intro_figure}, we conduct this analysis by reconstructing existing hallucination detection benchmarks with negated expressions and introducing NegHalu, a dataset in which hallucination labels are newly assigned to account for the effects of negation.

% In this paper, we present an initial exploration of how negated text influences hallucination detection in LLMs. We propose three open research questions and work towards answering them.

% \textbf{RQ1}. LLMs는 긍정형/부정형 텍스트에 따라 환각과 사실 여부를 구분하는데 차이를 보이는가?
% \textbf{RQ2}. What internal mechanisms contribute to detecting hallucination failures when processing negated text?
% \textbf{RQ3}. Can 추론과 지식의 편집이 improve LLMs' hallucination problems in the negated text?

% To answer the research questions, 우리는 환각 탐지 태스크를 기반으로 LLMs이 부정형 텍스트가 입력으로 주어지는 경우 문맥의 변화와 환각 여부의 판별할 수 있는지 explore한다. 이러한 분석을 conduct하기 위해 we reconstruct existing hallucination detection benchmarks into negated text and create NegHalu, a dataset with newly relabeled hallucination outcomes based on the effects of negation. NegHalu reveals changes in LLM performance across pre-negated and post-negated text and highlights the potential emergence of new hallucinations caused by negation.

% (3) In this paper, 우리는 explore whether LLMs can reliably maintain their hallucination detection capabilities when presented with negated text as context. 우리는 기존의 hallucination detection benchmarks를 부정형 텍스트로 변형하고 그에 따른 환각 여부를 새롭게 레이블링한 NegHalu를 제작했다. This reconstructed dataset은 pre-negated 텍스트와 post-negated text에 대해서 LLMs의 성능 변화와 negation으로 인한 새로운 환각의 발생 가능성을 드러낸다. 

Our experiments show that Llama-2-7B \citep{touvron2023llama}, Llama-3-8B \citep{llama3modelcard}, and Mistral-7B-v0.3 \citep{jiang2023mistral} exhibit performance degradation in 17 out of 18 post-negated hallucination detection cases. Furthermore, when the same content is expressed in both affirmative and negated forms, models tend to exhibit a bias toward classifying post-negated scenarios as hallucinations rather than faithful statements. We demonstrate that the influence of negated text on hallucination detection extends across multiple tasks, including question answering (QA), dialogue, summarization, and completion, indicating its task-agnostic. Moreover, this phenomenon is observed across a diverse range of domains, including 10 general topics, science, and autobiographies, suggesting its broad applicability. For in-depth analyses, we employ lens observation to examine the internal states of LLMs during hallucination detection. Furthermore, we examine approaches to alleviate the impact of negated text without resorting to unrealistic external modules or excessive parameter modifications, assessing the effectiveness of in-context learning \citep{brown2020language}, Chain-of-Thought (CoT) reasoning \citep{wei2022chain}, and knowledge editing \citep{fang2024alphaedit} as potential solutions.

% Our experiments show Llama-2-7B \citep{touvron2023llama}, Llama-3-8B \citep{llama3modelcard}, and Mistral-7B-v0.3 \citep{jiang2023mistral}이 18건의 post-negated halluncation detection tasks 중에서 17건에서 성능 하락을 나타냄. 또한, 동일한 내용을 긍정/부정형으로 표현하는 경우에 post-negated scenarios에서 사실보다 환각으로 판단하는 편향이 발생함. The impact of negated text on hallucination은 question answering (QA), Dialogue, Summarization, and Completion으로 task-agnostic하게 나타남. 더 나아가, 10 general topics, science, autobiographies의 다양한 도메인에서 발생 가능한 것으로 여겨짐. For in-depth analysis, we utilize a lens observation method to examine the internal states of LLMs during the hallucination detection process. Additionally, we investigate 외부 모듈이나 불필요한 학습을 통한 파라미터 변형 없이 the potential for mitigating the impact of negated text through in-context learning, Chain-of-Thought (CoT) reasoning \citep{wei2022chain}, and knowledge editing.

% (4) 더 심화된 분석을 위해서 우리는 환각 탐지 과정에서 LLMs의 interal state를 파악할 수 있는 lens observation method를 활용한다. 또한, 우리는 in-context learning, chain-of-thought, 그리고 knowledge editing을 통해 부정형 텍스트로 인한 새로운 환각에 대한 완화 가능성을 분석한다.

\section{Related Work}
\paragraph{Negated Text in LLMs}
Handling negated text has been a long-standing challenge in NLP. \citet{minsky1997negative} emphasized the importance of understanding negated expressions and meanings, highlighting the need to integrate negation into NLP systems. Building on this foundation, \citet{morante2011annotation} analyzed how negation operates within the text, providing the essential groundwork for subsequent studies. Further exploration aimed to understand the effects of negation on semantic structures and meaning \citep{van2016building,khandelwal2020negbert}. \citet{kassner2020negated} and \citet{hossain2022analysis} demonstrated that models frequently overestimate their confidence in predictions, leading to errors when processing negated inputs. \citet{arnaout2022uncommonsense} and \citet{chen2023say} revealed that negated knowledge can introduce biases into LLMs, further complicating their performance. \citet{truong2023language} observed that even as the model size increases, the ability to effectively handle negation does not necessarily improve. Moreover, \citet{ye2023assessing} found that negation can cause significant performance drops, even when advanced strategies like chain-of-thought reasoning are employed. Most recently, \citet{bhar2024strong} highlighted how negation can lead to unique types of hallucinations in tasks such as natural language inference.

\paragraph{Hallucination in LLMs} 
The issue of hallucination in LLMs has gained increasing importance as these models are applied to various NLP tasks. Hallucination occurs when the output generated by an LLM either lacks logical consistency with the input or contradicts real-world facts \citep{ji2023survey,huang2023survey}. This phenomenon has been observed across a range of tasks, including machine translation \citep{dale2023detecting,guerreiro-etal-2023-hallucinations}, summarization \citep{zhao2020reducing,choubey2023cape}, and dialogue generation \citep{ji2023rho}. It is particularly problematic in high-stakes domains that demand high reliability, such as law \citep{magesh2024hallucination}, medicine \citep{farquhar2024detecting}, and science \citep{dong2024bamboo}. To address hallucinations, recent research has advanced both detection and mitigation techniques \citep{ji2023survey,huang2023survey,zhang2023siren}. Detection strategies range from word-level and sentence-level analysis \citep{huang2023look,yang2023new} to self-verification via sampling \citep{manakul2023selfcheckgpt} and methods employing eigen-scores \citep{cheninside}. For mitigation, approaches include employing decoding strategies with contrasting layers \citep{chuang2023dola}, leveraging knowledge graph embeddings \citep{ji2023rho}, and fine-tuning model parameters based on data quality \citep{choubey2023cape}. In this paper, we explore the unanswered scope of hallucination and address the lack of research on how negated text affects hallucination phenomena in LLMs. We analyze LLM performance on pre-negated and post-negated statements to identify the underlying causes of performance shifts and investigate strategies for mitigating these effects.

% \section{Experimental Setup}
% We investigate whether LLMs can accurately understand negated contexts and correctly infer hallucination statuses. To this end, we construct the \textbf{NegHalu} dataset, which consists of datasets restructured with negated text and corresponding new labels. Using NegHalu, we evaluate the hallucination detection performance of various LLMs and perform deeper analyses through lens observation, in-context learning, Chain-of-Thought (CoT) reasoning \citep{wei2022chain}, and Knowledge Editing.

% We probe LLMs가 부정형 텍스트로 쓰여진 context를 이해하고 올바르게 추론하여 환각 여부를 제대로 탐지할 수 있는지. 우리는 부정형 텍스트로 재구성된 데이터셋인 NegHalu를 제작하고, 해당 데이터셋을 활용하여 각 LLM의 환각 탐지 성능을 평가하며, lens observation, in-context learning, Chain-of-Thought (CoT), 그리고 Knowledge Editing을 사용해 더 깊은 분석을 진행한다. 

\section{NegHalu}
\paragraph{Source Datasets}
% Table 1 - 사용한 데이터셋의 갯수와, post-negation 이후의 변동을 보여주는 테이블 필요
% (1) 우리는 3개의 환각 탐지 데이터셋을 활용함: HaluEval, BamBoo, and Selfcheckgpt wiki (각 데이터셋에 대한 설명)
% (2) From the HaluEval dataset, we employ 각각 1500개씩 examples for question answering (Halu-QA), knowledge-grounded Dialogue (Halu-Dialogue), and summarization (Halu-Sum), allowing the model to detect hallucinations based on the given context and knowledge.
% (3) BamBoo 데이터셋에서는 우리는 요약문에 환각된 정보가 포함되어 있는지를 판단하는 태스크인 AbsHallu 200개, 그리고 각 문장이 사실인지 여부를 판단하는 태스크인 SenHallu에서 200개를 사용함. 
% (4) SelfCheckGPT-WikiBio에서는 각 문장에 환각 여부가 레이블되어 있는 238명의 인물에 대해 GPT-3가 생성한 문단을 사용했다. 

We utilize three hallucination detection datasets: HaluEval \citep{li2023halueval}, BamBoo \citep{dong2024bamboo}, and SelfCheckGPT-WikiBio \citep{manakul2023selfcheckgpt}. Each dataset is selected for its relevance to evaluating hallucination phenomena across various tasks. For detailed statistics and descriptions of each dataset and its subsets, please refer to Appendix~\ref{appendix:dataset_details}.

\begin{table*}[t]
\footnotesize
\centering
\resizebox{0.9\textwidth}{!}{
\begin{tabular}{p{\textwidth}}
    \toprule
    \#\#\# \textbf{Restructuring Task Prompt | Hallucination Detection Datasets (Round 1)} \\ 
    \rowcolor{headercolor} \textbf{\#\#System}: \\
    \rowcolor{conceptcolor} You are a human annotator and an English native speaker, restructuring text according to given instructions. \\ 
    \rowcolor{headercolor} \textbf{\#\#Instruction}: \\ 
    \rowcolor{conceptcolor} - The provided text is structured as [A], [B], [C], and [LABEL]. \\ 
    \rowcolor{conceptcolor} - [C] includes content that allows for determining the presence of hallucination based on [A] and [B]. \\ 
    \rowcolor{conceptcolor} - [LABEL] indicates whether [C] is ``Hallucinated" if it contains hallucinations, or ``True" if it does not. \\ 
    \rowcolor{conceptcolor} Your task is to restructure [C] into a negative statement [NEW C] by adding ``not" ONCE so that [Label] changes accordingly. \\ 
    \rowcolor{headercolor} \textbf{\#\#Requirements}: \\
    \rowcolor{conceptcolor} - Do not use double negatives. \\ 
    \rowcolor{conceptcolor} - Adding ``not" only once is mandatory. \\ 
    \rowcolor{conceptcolor} - The final result MUST align with real-world facts and commonsense. \\ 
    \rowcolor{conceptcolor} - Generate only the text within [NEW C], omitting any other content. \\ 
    \rowcolor{headercolor} \textbf{\#\#Input Format}: \\
    \texttt{[A]: \{GIVEN A TEXT\}} \\ 
    \texttt{[B]: \{GIVEN B TEXT\}} \\ 
    \texttt{[C]: \{GIVEN C TEXT\}} \\ 
    \texttt{[LABEL]: \{HALLUCINATED OR TRUE\}} \\ 
    \texttt{[NEW C]}: \\ 
    \texttt{[NEW LABEL]}: \\ 
    \bottomrule
\end{tabular}}
\caption{Generalized prompt used for hallucination detection datasets to restructure new negated texts and labels.}
\label{tab:table1}
\end{table*}

\begin{table}[t]
\footnotesize
\centering
\resizebox{0.9\linewidth}{!}{
\begin{tabular}{p{\linewidth}}
    \toprule
    \#\#\# \textbf{Evaluation Task Prompt | Data Verification (Round 2)} \\
    \rowcolor{headercolor} \textbf{\#\#System}: \\
    \rowcolor{conceptcolor} You are a meticulous evaluator, carefully assessing if generated responses meet specific instructions and requirements. \\
    \rowcolor{headercolor} \textbf{\#\#Evaluation Instructions}: \\
    \rowcolor{conceptcolor} - The text provided is structured as [A], [B], [C], [NEW C], [LABEL], and [NEW LABEL]. \\
    \rowcolor{conceptcolor} - Your task is to evaluate the [NEW C] and [NEW LABEL] for the following criteria: \\
    \rowcolor{conceptcolor}   1. \textbf{Logical Negation}: Ensure that [NEW C] negates [C] logically to change the meaning and the [LABEL] appropriately. \\
    \rowcolor{conceptcolor}   2. \textbf{New Label Validity}: Check that [NEW C] is appropriate for the assigned [NEW LABEL]. \\
    \rowcolor{headercolor} \textbf{\#\#Output Format for Evaluation}: \\
    \rowcolor{conceptcolor} After evaluating each criterion, rate it as "Pass" or "Fail." If a criterion fails, provide a brief reason. The final output should use the following format: \\
    \rowcolor{headercolor} \textbf{\#\#Evaluation Criteria:} \\
    \rowcolor{conceptcolor} - Logical Negation: Pass / Fail \\
    \rowcolor{conceptcolor} - NEW Label Validity: Pass / Fail \\
    \rowcolor{headercolor} \textbf{\#\#Output Format}: \\
    \texttt{[RESULT]: [Pass, Pass]}  \\
    \rowcolor{headercolor} \textbf{\#\#Input Format}: \\
    \texttt{[RESULT]:} \\
    \bottomrule
\end{tabular}}
\caption{Generalized prompt for evaluating logical negations and the validity of new negated text and labels.}
\label{tab:table2}
\end{table}

\paragraph{Post Negation}
To analyze the impact of negated text on hallucination detection, we introduce a post-negation transformation applied to key fields in each dataset that are crucial for determining the presence of hallucination. This transformation compels the model to re-evaluate its predictions in the context of post-negated input. The following fields are transformed into their post-negated text: the \texttt{`answer'} field for Halu-QA, the \texttt{`response'} field for Halu-Dialogue, the \texttt{`summary} field for Halu-Sum, the \texttt{`hypothesis'} field for AbsHallu and SenHallu, and the \texttt{`generated text'} field for SelfCheckGPT-WikiBio. Table \ref{tab:table1} illustrates the prompt templates used to generate post-negated texts and corresponding new labels. To create the negated versions of the datasets, we utilize the GPT-4 omni (\texttt{gpt-4o-2024-08-06}) \citep{openai2023gpt4} API in a two-round process. 

As described in Table~\ref{tab:table1}, in the first round (\textbf{Round 1}), we instruct the model to generate post-negated texts based on the given context and knowledge, aiming to maintain logical consistency independently. In this setting, “logical consistency” means that the insertion of a negation marker should transform the original pre-negated text, whether it is hallucinated or factual, so that the resulting meaning coherently aligns with the new label. This requirement is explicitly stated in the instruction: “\textit{Your task is to restructure [C] into a negative statement [NEW C] by adding ‘not’ ONCE so that [Label] changes accordingly},” which is intended to ensure that the negated text both reverses the original meaning and remains consistent with the assigned label.
Additionally, new labels are systematically assigned by considering the changes in the context of pre-negated examples, a process that requires modifying the original hallucination statuses and corresponding labels. This approach is designed to promote a definitive change in the labels of post-negated texts and to maximize the generation of examples in which the insertion of negation results in a logically valid label shift.

\paragraph{Data Verification}
To validate the plausibility of the post-negated text and the corresponding new labels, we conduct a second verification round (\textbf{Round 2}). In this step, three GPT-4 omni models with different temperatures (0, 0.7, 1.2) independently evaluate the outputs from Round 1. The evaluation prompt, shown in Table~\ref{tab:table2}, is designed to control and assess data quality. Each example is evaluated on two criteria: \textbf{Logical Negation}, whether the post-negated text logically and effectively transforms the meaning of the pre-negated text, and \textbf{New Label Validity}, whether the revised text is correctly assigned the appropriate hallucination label. For each criterion, all three evaluators provide a binary judgment (\texttt{Pass} or \texttt{Fail}), yielding an outcome such as [Pass, Pass] if both aspects are satisfied. Only examples that unanimously receive "Pass" for both criteria are retained in the NegHalu dataset. This stringent process, described in more detail with qualitative examples in Appendix~\ref{sec:verfication}, demonstrates the robustness and reliability of our logical transformation and label assignment. As a result, the final NegHalu dataset consists exclusively of high-quality, validated negated examples.

\paragraph{Qualitative Analysis}
\label{sec:post-negation}
As shown in Table~\ref{tab:appendix_negation_qualitative}, minimal negation operations in HaluEval effectively flip factual status: inserting “not” into a true claim about FC Bayern Munich introduces a hallucination, while adding “did not” to an erroneous claim about Michael Sheen corrects it. Similarly, in Table~\ref{tab:appendix_bamboo_qualitative}, Bamboo examples show that accurate hypotheses (e.g., the competitiveness of translated data models) become false once negated, while hallucinated statements about summary generation methods are corrected through the same operation.

In Table~\ref{tab:appendix_selfcheck_qualitative}, SelfCheckGPT-WikiBio further demonstrates this precision, where a true claim about Lee Hsien Loong is rendered false by negation, while a false claim about Admiral Aylmer is corrected. These cases across dialogue, QA, and summarization confirm that our negation strategy reliably induces or removes hallucinations with minimal intervention, ensuring robustness and low noise across datasets.

\paragraph{Human Evaluation}
In addition to automated verification, the authors manually inspected all 1,950 generated examples to ensure the overall plausibility and factual consistency of the NegHalu dataset. Human evaluation served as a precautionary step to prevent the inclusion of nonsensical or factually incorrect content and to ensure alignment with real-world knowledge and commonsense. For example, if a question asks, “\textit{Which industry do Richard Hawley and Chicago's Catherine belong to?}” and the original hallucinated answer is “\textit{Richard Hawley is a chef},” then the negated output “\textit{Richard Hawley is not a chef}” would contradict known facts about Richard Hawley. As a result of this manual review, we identified and revised 13 nonsensical or factually incorrect examples in HaluEval, 11 in SelfCheckGPT-WikiBio, and 3 in BamBoo. Additionally, for BamBoo–SenHallu, 2 cases involving double negation were rephrased as single negation to maintain contextual clarity.

\begin{table}[t]
\footnotesize
\centering
\resizebox{1\linewidth}{!}{
\begin{tabular}{lcc}
    \toprule
    \textbf{Dataset} & \textbf{Post Negation (Pre/Post)} & \textbf{After Verification (Pre/Post)} \\
    \midrule
    \textbf{NegHalu} & 6,257 / 6,257 & 1,950 / 1,950 \\
    \midrule
    $\vdash$\texttt{HaluEval - QA} & 1,500 / 1,500 & 400 / 400 \\
    $\vdash$\texttt{HaluEval - Dialogue} & 1,500 / 1,500 & 400 / 400 \\
    $\vdash$\texttt{HaluEval - Sum} & 1,500 / 1,500 & 400 / 400 \\
    $\vdash$\texttt{BamBoo - AbsHallu} & 200 / 200 & 152 / 152 \\
    $\vdash$\texttt{BamBoo - SenHallu} & 200 / 200 & 136 / 136 \\
    $\vdash$\texttt{SelfCheckGPT-WikiBio} & 1,357 / 1,357 & 462 / 462 \\
    \bottomrule
\end{tabular}}
\caption{Dataset Overview for NegHalu and Its Subsets. SelfCheckGPT-WikiBio represents the number of sentences obtained by splitting 238 paragraphs.}
\label{tab:table3}
\end{table}

\paragraph{Data Statistics and Label Balancing}
Table \ref{tab:table3} presents the statistics for NegHalu, summarizing the results after the two-round process of generation and verification. The NegHalu dataset consists of 1,950 samples, reconstructed from the three original hallucination detection datasets with post-negated text and corresponding new labels. Each subset within NegHalu preserves the evaluation framework and methodology of its respective source benchmark. To mitigate label imbalances, we maintain an equal ratio of hallucinated to faithful samples in HaluEval and SelfCheckGPT-WikiBio, while BamBoo is excluded from this adjustment because of its comparatively smaller size. Additionally, during the data verification process, we include only post-negated examples where the meaning of the original text has been altered, resulting in a label change. This ensures that the ratio of labels between pre- and post-negated examples remains equal in the final dataset. This balanced distribution makes it easier to analyze the model's preference for specific labels when presented with pre- or post-negated examples.

% Post-negated text와 new label의 타당성을 검증하기 위해, 우리는 Round 2에서 GPT-4o가 Round1에서 생성한 결과물을 스스로 평가하도록 했다. Table 2은 data curation을 통해 생성한 데이터의 품질을 control하기 위한 평가 프롬프트를 나타낸다. As shown in Table 2, 우리는 post-negated text가 pre-negated text의 의미를 논리적으로 타당하게 변형하였는지, 그리고 new label에 타당한 내용인지 평가한다. 2개의 평가 기준인 Logical Negation과 NEW Label validity는 Pass or Fail로 각 샘플의 적합성을 채점한다. 우리는 2개의 평가 기준에서 모두 PASS를 받은 샘플만을 최종 NegHalu 데이터셋으로 사용한다. 

% Table 3은 2개의 Round를 통해 제작하고 필터링한 NegHalu의 통계를 나타낸다. NegHalu는 기존 3개의 환각 탐지 데이터셋을 부정형 텍스트로 재구성하고 새로운 레이블을 부여한 2,412개의 샘플로 구성된다. NegHalu의 각 데이터셋은 기존 벤치마크의 평가 방식을 동일하게 유지한다. 또한, 우리는 source dataset의 갯수가 상대적으로 적었던 Bamboo를 제외한 HaluEval과 SelfCheckGPT-WikiBio에 대해서 환각과 사실의 비율이 동일하도록 구성했다. 

\section{Experimental Setup}
\paragraph{Models}
To achieve diverse and representative coverage in our experiments, we focus on selecting LLMs that are well-regarded in the open-source community and frequently serve as benchmarks in follow-up research. Our experimental framework is designed to investigate whether these models exhibit consistent patterns or distinctive behaviors. The models chosen for this study—Llama2 \citep{touvron2023llama} (\textit{meta-llama/Llama-2-7b-chat-hf}), Llama3 \citep{llama3modelcard} (\textit{meta-llama/Meta-Llama-3-8B-Instruct}), Mistral \citep{jiang2023mistral} (\textit{mistralai/Mistral-7B-Instruct-v0.3}), and Qwen3 \citep{bai2023qwen} (\textit{Qwen/Qwen3-4B})—are known for their ability to generate high-quality outputs from provided instructions and for their adherence to the exact match metric, making them well-suited for our evaluation tasks.

\paragraph{Datasets and Evaluation}
We adopt standard hallucination detection benchmarks: HaluEval, BamBoo, and SelfCheckGPT-WikiBio. Evaluations use pairs of pre-negated and post-negated examples, with original and updated labels, under consistent settings. \textbf{HaluEval} and \textbf{BamBoo} use binary classification to assess whether model outputs contain hallucinations, reporting Accuracy (HaluEval) and F1 score (BamBoo). \textbf{SelfCheckGPT-WikiBio} measures sentence-level accuracy against human annotations, treating any inaccuracy as hallucination. For all datasets, we compare model performance before and after negation to analyze label shifts.

% SelfCheckGPT-WikiBio는 GPT-3가 생성한 문단의 각 문장에 대해서 모델이 inaccurate라 판단하는 경우 1, accurate라 판단하는 경우 0을 생성하도록 한다\footnote{우리는 평가의 모호함을 방지하기 위해 minor inaccurate로 분류된 문장도 강한 환각으로 분류}. 모델의 response는 human annotated judgments와 비교한 accuracy로 평가한다. 우리는 부정형 텍스트와의 더 명확한 비교를 위해 모델의 응답에 대한 누적 점수와 레이블과의 절대적인 거리를 추가적으로 비교한다. 

% \p{Analysis Methods}
% \label{sec:analysis}
% Based on the evaluation results of NegHalu, we aim to trace the causes of additional hallucinations induced by negated text and explore applicable mitigation strategies. To conduct a more in-depth analysis, we employ the following four methods:

\paragraph{Lens Observation}
This method examines intermediate representations and attention distributions within the model to uncover how negated text influences the generation process. We employ a lens observation method to analyze LLMs' processing of pre- and post-negated text across layers. The Logit Lens \citep{nostalgebraist2020,dar2023analyzing} projects internal model states onto the vocabulary space, tracking prediction changes layer by layer. Lens observation helps pinpoint token-wise changes across layers, offering a comprehensive understanding of how negation affects predictions. We focus on observing the final token of the input and the first token generated by the model to measure the judgment in response to negated input.

%We analyze how negated text affects model predictions using Logit Lens \citep{nostalgebraist2020,dar2023analyzing}, which projects internal states onto the vocabulary space layer by layer. By observing token-wise changes—especially at the final input token and the model’s first generated token—we gain insight into how negation influences hallucination judgments.

\paragraph{In-Context Learning} We assess model adaptability to negation by providing pre- and post-negated examples together in the input context \citep{brown2020language,dong2024survey}. Our experiments include zero-shot, two-shot, and four-shot settings, each offering different combinations of faithful and hallucinated examples as shown in Table~\ref{tab:table4}. The two-shot setting serves as our default.

% \paragraph{Chain-of-Thought} We employ step-by-step reasoning to examine whether it can reduce hallucinations in scenarios involving negated text \citep{wei2022chain}. By analyzing the intermediate reasoning steps, we aim to understand the rationale behind hallucination classification for pre- and post-negated examples and improve the model's performance. In our experiments, models are prompted to first articulate their reasoning process before determining whether the input contains hallucinations. Table \ref{tab:table4} provides a generalized example of the CoT prompt templates applied across the datasets.

\paragraph{Chain-of-Thought} We use step-by-step reasoning to evaluate whether it reduces hallucinations with negated text \citep{wei2022chain}. Models are prompted to explain their reasoning before making hallucination judgments. Table~\ref{tab:table4} shows the CoT prompt template.

%We use step-by-step reasoning (Chain-of-Thought) to assess its effect on hallucination detection with negated text \citep{wei2022chain}. Models are prompted to explain their reasoning before classifying each example. See Table~\ref{tab:table4} for the prompt template.

\paragraph{Knowledge Editing}
In our experiments, we adopt the approach of AlphaEdit \citep{fang2024alphaedit} to examine and address hallucination issues in negated text scenarios. To prevent disruptions to parametric knowledge, AlphaEdit projects parameter updates onto the null space of the preserved knowledge, ensuring minimal interference with existing factual associations. Our study focuses on using the null-space constraint to preserve affirmative knowledge while updating negated knowledge. We approximate the covariance matrix of preserved knowledge and use causal tracing to identify layers for editing. By leveraging 100,000 (subject, relation, object) triplets from Wikipedia \citep{meng2023massediting}, we construct a basis for preserved knowledge and target layers critical for encoding it, ensuring edits address negated knowledge without unwanted parameter updates. Training details are provided in Appendix \ref{sec:appendix}.
%In our experiments, we use AlphaEdit \citep{fang2024alphaedit} to address hallucination in negated text. AlphaEdit constrains parameter updates to the null space of preserved (affirmative) knowledge, minimizing interference with existing facts. We identify edit layers using causal tracing and construct the knowledge basis from 100{,}000 Wikipedia triplets \citep{meng2023massediting}. Additional training details are in Appendix~\ref{sec:appendix}.

\begin{table*}[h]
\centering
\resizebox{1\textwidth}{!}{
    \renewcommand{\arraystretch}{1.2}
    \begin{tabular}{ccccccccccc}
\toprule
\multirow{2}{*}{\textbf{Models}} & \multicolumn{2}{c}{\textbf{HaluEval-QA (Acc)}} & \multicolumn{2}{c}{\textbf{HaluEval-Dialogue (Acc)}} & \multicolumn{2}{c}{\textbf{HaluEval-Sum (Acc)}} & \multicolumn{2}{c}{\textbf{BamBoo-AbsHallu (P/R/F1)}} & \multicolumn{2}{c}{\textbf{BamBoo-SenHallu (P/R/F1)}} \\
\cmidrule(lr){2-3} \cmidrule(lr){4-5} \cmidrule(lr){6-7} \cmidrule(lr){8-9} \cmidrule(lr){10-11}
& \textbf{Pre} & \textbf{Post} & \textbf{Pre} & \textbf{Post} & \textbf{Pre} & \textbf{Post} & \textbf{Pre} & \textbf{Post} & \textbf{Pre} & \textbf{Post}  \\
\midrule
Llama-2-7B & 0.4825 & \textbf{0.4950} & \textbf{0.6150} & 0.4975 & \textbf{0.4750} & 0.4625 & \textbf{59.8/73.6/66.0} & 38.4/54.1/44.9 & \textbf{68.7/77.3/72.7} & 34.3/47.9/40.0 \\
Llama-3-8B & \textbf{0.7650} & 0.5525 & \textbf{0.7825} & 0.4500 & \textbf{0.6600} & 0.5175 & \textbf{59.5/96.7/72.7} & 38.9/80.3/52.4 & \textbf{69.9/97.7/81.5} & 31.1/58.3/40.6\\
Mistral-7B-v0.3 & \textbf{0.5900} & 0.5200 & \textbf{0.7050} & 0.5100 & \textbf{0.5950} & 0.5125 & \textbf{61.1/100/75.8} & 32.6/49.2/39.2 & \textbf{77.3/96.6/75.9} & 31.1/47.9/37.7 \\
Qwen3-4B & \textbf{0.4425} & 0.2775 & \textbf{0.7425} & 0.5150 & \textbf{0.5625} & 0.4675 & \textbf{64.5/100/78.4} & 50.0/21.3/29.9 & \textbf{82.8/93.2/87.7} & 46.7/14.6/22.2 \\

\bottomrule
\end{tabular}}
\caption{Performance comparison of models across HaluEval and Bamboo subsets in the NegHalu. \textbf{Acc} represents accuracy, and \textbf{P/R/F1} denotes precision, recall, and F1-score. Bold text indicates the higher performance between \textbf{Pre}- and \textbf{Post}-negated scenarios for the input example.}
\label{tab:table5}
\end{table*}

\begin{table*}[h]
\centering
\resizebox{1\textwidth}{!}{
    \renewcommand{\arraystretch}{1.2}
    \begin{tabular}{cccccccccc}
\toprule
\multirow{2}{*}{\textbf{Correct Answers}} & \multicolumn{2}{c}{\textbf{HaluEval (Pre-negated)}} & \multicolumn{2}{c}{\textbf{HaluEval (Post-negated)}} & \multicolumn{2}{c}{\textbf{BamBoo (Pre-negated)}} & \multicolumn{2}{c}{\textbf{BamBoo (Post-negated)}} \\
\cmidrule(lr){2-3} \cmidrule(lr){4-5} \cmidrule(lr){6-7} \cmidrule(lr){8-9}
& \textbf{Halu. = "YES"} & \textbf{Halu. = "NO"} & \textbf{Halu. = "YES"} & \textbf{Halu. = "NO"} &\textbf{Halu. = "YES"} & \textbf{Halu. = "NO"} & \textbf{Halu. = "YES"} & \textbf{Halu. = "NO"} \\
\midrule
Llama-2-7B & 408 & 289 & 504 $\bigtriangleup$ & 78 $\bigtriangledown$ & 33 & 135 & 82 $\bigtriangleup$ & 56 $\bigtriangledown$\\
Llama-3-8B & 398 & 443 & 422 $\bigtriangleup$ & 186 $\bigtriangledown$ & 12 & 174 & 40 $\bigtriangleup$ & 77 $\bigtriangledown$\\
Mistral-7B-v0.3 & 413 & 427 & 578 $\bigtriangleup$ & 39 $\bigtriangledown$ & 26 & 176 & 66 $\bigtriangleup$ & 53 $\bigtriangledown$\\
Qwen3-4B & 410 & 191 & 467 $\bigtriangleup$ & 37 $\bigtriangledown$ & 42 & 173 & 158 $\bigtriangleup$ & 20 $\bigtriangledown$\\
\bottomrule
\end{tabular}}
\caption{Label distribution across pre- and post-negated scenarios in the HaluEval and BamBoo subsets of the NegHalu. \textbf{Halu.} represents the number of examples classified as "Hallucinated = YES" or "Hallucinated = NO" among correctly predicted answers for each model. $\bigtriangleup$ and $\bigtriangledown$ indicate increases and decreases, respectively, in Post-negated compared to Pre-negated.}
\label{tab:table6}
\end{table*}

\begin{table}[h]
\centering
\resizebox{1\linewidth}{!}{
    \renewcommand{\arraystretch}{1.2}
    \begin{tabular}{ccccccccc}
\toprule
\multirow{2}{*}{\textbf{Models}} & \multicolumn{6}{c}{\textbf{SelfCheckGPT-WikiBio}} \\
\cmidrule(lr){2-3} \cmidrule(lr){4-5} \cmidrule(lr){6-7} 
& \textbf{Pre (Acc) } & \textbf{Post (Acc)} & \textbf{Pre (Avg)} & \textbf{Post (Avg) } & \textbf{Pre (Dis)} & \textbf{Post (Dis)} \\
\midrule
Llama-2-7B & \textbf{0.6169} & 0.4686 & 0.5357 & \underline{0.9205} & 0.3810 & \underline{0.5319} \\
Llama-3-8B & \textbf{0.6786} & 0.4610 & 0.3598 & \underline{0.8853} & 0.3209 & \underline{0.5390} \\
Mistral-7B-v0.3 & \textbf{0.6450} & 0.4881 & 0.2489 & \underline{0.6245} & 0.3550 & \underline{0.5119} \\
Qwen3-4B & \textbf{0.5866} & 0.5022 & 0.8939 & \underline{0.9892} & 0.4134 & \underline{0.4978} \\

\bottomrule
\end{tabular}}
\caption{Performance comparison of models on the SelfCheckGPT-WikiBio subset in the NegHalu. \textbf{Acc} represents accuracy, \textbf{Avg} denotes the average of cumulative scores for model responses (0: True, 1: Hallucinated), and \textbf{Dis} indicates the absolute distance between the assigned labels and responses. Bold and Underline highlight the higher scores based on each column.}
\label{tab:table7}
\end{table}

\section{Results}
In this section, we present experimental findings and analyses that provide answers to the research questions raised in this study. A detailed summary of the answers to the research questions is provided in the Appendix \ref{sec:appendix_rq}.

\paragraph{A1. LLMs Exhibit Degradation and Bias in Hallucination Detection for Negated Text}

\paragraph{NegHalu - HaluEval}
Table \ref{tab:table5} presents the performance of models on the QA, Dialogue, and Summarization tasks in the HaluEval dataset under pre- and post-negated input scenarios. Across all tasks, model performance generally decreases when detecting hallucinations with post-negated inputs. Interestingly, Llama2 demonstrates robustness to negated inputs, maintaining performance levels comparable to non-negated scenarios, while Llama3 and Mistral experience significant performance drops. Qwen3-4B shows mixed behavior, performing well on Dialogue and Summarization in the pre-negated setting but dropping sharply in QA and post-negated cases. As shown in Table \ref{tab:table6}, models exhibit a strong tendency to classify negated texts as hallucinated, with an 21.0\% increase in hallucination predictions and a 74.8\% decrease in faithfulness judgments. This suggests that models may develop biases toward specific labels when processing negated inputs.

% Table \ref{tab:table5}는 QA Dialogue Summarization task에서 pre- and post-negated 입력 시나리오에 따른 성능을 보인다. post-negated 예시를 입력으로 하여 환각을 탐지하는 경우 전반적으로 모델의 성능이 감소한다. 흥미롭게도 Llama2는 기존 성능에 비해 부정형 입력에 대해 강인한 모습을 보이며, Llama3와 Mistral은 큰 폭으로 성능 하락을 보인다. As described in Table \ref{tab:table6}, 모델들은 부정형 텍스트를 환각이 포함되었다고 판단하는 경향이 강하며, 대략 18.71\% 더 높은 비율로 환각이 있는 것으로 판단하고, faithfulness하다고 판단할 비율이 약 67.14\% 감소함. 이는 부정형 텍스트에 대해서 모델이 특정 레이블에 편향되는 현상을 암시함.

\paragraph{NegHalu - BamBoo}
Table \ref{tab:table5} compares the performance of models in the BamBoo dataset for AbsHallu and SenHallu tasks under pre- and post-negated input scenarios. Across all models and metrics, the post-negated scenario consistently results in significant performance declines, with decreases as large as 51.8. Furthermore, Table \ref{tab:table6} highlights greater label distribution shifts in BamBoo compared to HaluEval. For post-negated inputs, hallucination predictions increase by approximately 206.2\%, while faithfulness judgments decrease by around 68.7\%. These findings align with the trends observed in HaluEval, further confirming that LLMs face considerable challenges in detecting hallucinations when processing negated text.

% Table \ref{tab:table5}는 BamBoo의 AbsHallu와 SenHallu 태스크에서 pre- and post-negated 입력 시나리오에 따른 성능을 비교한다. post-negated 시나리오는 pre-negated에 비해서 모든 metric과 모델에서 큰 성능 하락을 보이며, 최대 51.8의 감소를 보이기도 한다. Table \ref{tab:table6}는 BamBoo에서 HaluEval보다 정답으로 선택한 레이블 distribution에 있어서 더 큰 변동을 보인다. 부정형 입력에 대해서는 환각으로 판단할 비율이 약 164.79\% 상승하고, faithful하다고 판단할 비율이 약 61.65\% 감소한다. 이러한 결과는 HaluEval과 상당히 유사한 경향을 나타내며, LLMs가 부정형 텍스가 입력으로 주어지는 경우 환각 탐지에 있어서 큰 어려움을 겪고 있음을 검증한다.

\paragraph{NegHalu - SelfCheckGPT}
Table \ref{tab:table7} shows the performance of models in classifying hallucinations on a sentence-level basis within SelfCheckGPT-WikiBio under pre- and post-negated input scenarios. Consistent with earlier datasets, model performance consistently decreases with negated inputs. Additionally, the proportion of hallucination judgments increases across all models, accompanied by growing gaps between predictions and ground-truth labels. These results suggest that negated text inputs introduce new and unintended hallucination patterns, reflecting similar trends across all evaluated datasets.

% Table \ref{tab:table7}는 SelfCheckGPT-WikiBio에 대한 pre- and post-negated 입력 시나리오에 따른 각 문장별로 환각 여부를 판별한 성능을 나타냄. 앞선 데이터셋들의 실험 결과와 동일하게 각 모델에 대한 성능은 부정형 입력에 대해서 일관되게 감소한다. 또한, 환각으로 판단하는 비율이 모든 모델에서 높아지며, 정답 레이블과의 점수 격차도 커지면서 환각 여부에 대한 판단 능력이 크게 감소한다. 부정형 텍스트에 대한 실험 결과는 모든 데이터셋에서 유사한 경향성을 보이며, 의도하지 않은 새로운 환각의 발생 가능성을 암시함.

\begin{figure*}[h]
\centering
\includegraphics[width=1\linewidth]{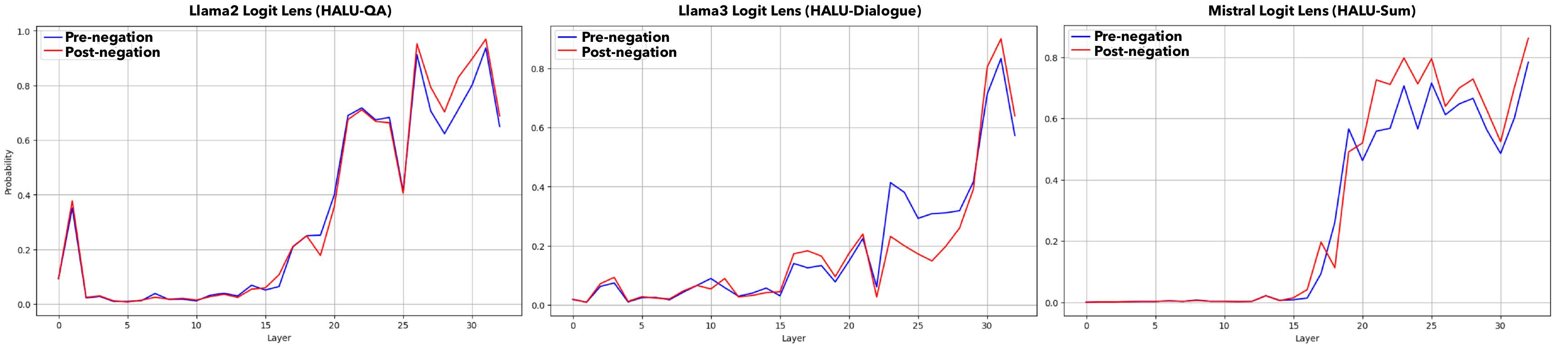}
\caption{Logit lens results showing probability shifts for pre- and post-negated examples in the HaluEval subsets of NegHalu. Each curve tracks the probability of the first output token across the layers of the model, comparing pre-negated and post-negated inputs. The \textcolor{blue}{blue curves} represent scenarios with pre-negated inputs, while the \textcolor{red}{red curves} indicate scenarios with post-negated inputs.}
 \label{fig:figure1}
\end{figure*}

\begin{table*}[h]
\centering
\resizebox{1\textwidth}{!}{
    \renewcommand{\arraystretch}{1.2}
    \begin{tabular}{cccccccccc}
\toprule
\multirow{2}{*}{\textbf{Models}} & \multicolumn{3}{c}{\textbf{HaluEval (Pre/Post) - Accuracy}} & \multicolumn{3}{c}{\textbf{BamBoo (Pre/Post) - F1}} & \multicolumn{3}{c}{\textbf{SelfCheckGPT (Pre/Post) - Accuracy}} \\
\cmidrule(lr){2-4} \cmidrule(lr){5-7} \cmidrule(lr){8-10}
& \textbf{0-shot} & \textbf{2-shot} & \textbf{4-shot} & \textbf{0-shot} & \textbf{2-shot} & \textbf{4-shot} & \textbf{0-shot} & \textbf{2-shot} & \textbf{4-shot} \\
\midrule
Llama-2-7B & 0.4466 / 0.4500 & 0.5241 / 0.4850 & 0.4975 / 0.4325 & 73.2 / 47.3 & 69.4 / 42.5 & 63.6 / 43.4 & 0.5519 / 0.4838 & 0.6169 / 0.4686 & 0.5758 / 0.4805 \\
Llama-3-8B & 0.0233 / 0.0025 & 0.7358 / 0.5066 & 0.7333 / 0.5083 & 77.4 / 43.9 & 77.1 / 46.5 & 78.1 / 48.1 & 0.6851 / 0.4729 & 0.6786 / 0.4610 & 0.6742 / 0.4729 \\
Mistral-7B-v0.3 & 0.5600 / 0.5100 & 0.6300 / 0.5142 & 0.5167 / 0.5058 & 80.9 / 36.8 & 75.9 / 38.5 & 81.1 / 36.2 & 0.6515 / 0.4686 & 0.6450 / 0.4881 & 0.5703 / 0.4946 \\
Qwen3-4B & 0.5317 / 0.3867 & 0.5825 / 0.4200 & 0.5817 / 0.4750 & 82.8 / 28.2 & 83.1 / 26.1 & 83.6 / 39.8 & 0.5779 / 0.4232 & 0.5866 / 0.5022 & 0.5325 / 0.4989 \\
\bottomrule
\end{tabular}}
\caption{Performance comparison of models across HaluEval, BamBoo, and SelfCheckGPT subsets in the NegHalu dataset. \textbf{Accuracy} and \textbf{F1} scores are reported for pre- and post-negated scenarios across 0-shot, 2-shot, and 4-shot.}
\label{tab:table8}
\end{table*}

\begin{table*}[h]
\centering
\resizebox{0.8\textwidth}{!}{
    \renewcommand{\arraystretch}{1.2}
    \begin{tabular}{cccccccc}
\toprule
\multirow{2}{*}{\textbf{Models}} & \multicolumn{2}{c}{\textbf{HaluEval (Pre/Post) - Acc}} & \multicolumn{2}{c}{\textbf{BamBoo (Pre/Post) - F1}} & \multicolumn{2}{c}{\textbf{SelfCheckGPT (Pre/Post) - Acc}} \\
\cmidrule(lr){2-3} \cmidrule(lr){4-5} \cmidrule(lr){6-7}
& \textbf{2-shot + CoT} & \textbf{4-shot + CoT} & \textbf{2-shot + CoT} & \textbf{4-shot + CoT} & \textbf{2-shot + CoT} & \textbf{4-shot + CoT} \\
\midrule
Llama-2-7B & 0.5308 / 0.4850 & 0.5808 / 0.4617 & 48.3 / 40.0 & 40.0 / 33.5 & 0.5108 / 0.5000 & 0.5043 / 0.5000 \\
Llama-3-8B & 0.7388 / 0.5067 & 0.6992 / 0.5108 & 69.0 / 44.3 & 72.6 / 42.2 & 0.5130 / 0.5000 & 0.5020 / 0.5000 \\
Mistral-7B-v0.3 & 0.5400 / 0.5147 & 0.7000 / 0.5192 & 80.2 / 38.6 & 80.3 / 37.8 & 0.6082 / 0.4805 & 0.5610 / 0.5065 \\
Qwen3-4B & 0.5767 / 0.5333 & 0.5667 / 0.5350 & 82.1 / 26.7 & 81.8 / 39.2 & 0.6494 / 0.5000 & 0.5974 / 0.5011 \\
\bottomrule
\end{tabular}}
\caption{Performance comparison of models across HaluEval, BamBoo, and SelfCheckGPT subsets in the NegHalu dataset. \textbf{Accuracy (Acc)} and \textbf{F1} scores are reported for pre- and post-negated scenarios across CoT conditions.}
\label{tab:table9}
\end{table*}

\paragraph{A2. Lack of Distinction Between Pre- and Post-Negated Text}

\paragraph{Logit Lens Observation} When negated text is provided as input, hallucination detection performance decreases, and models show a bias toward classifying the input as containing hallucinations. To understand the underlying cause, we trace the internal states of LLMs during the hallucination detection. We observe the final token of inputs containing pre- and post-negated text. Figure \ref{fig:figure1} illustrates the probability shifts for the next token by analyzing the hidden states at each layer of Llama2, Llama3, and Mistral on NegHalu. While the specific layers and magnitude of these shifts vary across models, strong probability fluctuations generally occur in the middle layers, followed by another significant fluctuation near the final layers. 
% This pattern suggests that the models' internal representations undergo substantial transformations in response to negated inputs, contributing to the observed biases in hallucination detection~\citep{cheninside,jiang2024large}.

\paragraph{Subtle Differences} LLMs in Figure \ref{fig:figure1} show only marginal differences between pre- and post-negated examples. This implies that despite the transformation of context or knowledge induced by negation, the models fail to clearly recognize these changes when determining hallucination. This phenomenon appears to be associated with the treatment of negation within the model's latent representation, where negation functions more as a single token rather than as a logical operator \citep{bhar2024strong}. Moreover, post-negated examples generally exhibit greater confidence in their decisions or show significant fluctuations near the final layers. This behavior closely resembles the token probability shifts observed in hallucination-inducing cases, as reported in \citep{cheninside,jiang2024large}, and is considered a contributing factor to the bias toward hallucination judgments for negated examples.

% 부정형 텍스트가 입력으로 주어지는 경우에 환각 탐지 성능이 하락하며, 환각이 존재하는 것으로 편향된 판단을 하는 원인은 파악하기 위해, 우리는 trace LLMs' internal state 환각 판별 과정. Based on the setup in $\S$\ref{sec:analysis}, 우리는 pre- and post-negated text를 포함한 입력의 마지막 토큰을 logit lens를 활용하여 observe한다. Figure 1 shows the probability shift for the next token by examining the hidden states at each layer for Llama2, Llama3, and Mistral on NegHalu. 모델마다 구체적인 위치와 정도는 다르지만, 전반적으로 중간 레이어 사이에서 강력한 확률 변동이 나타나며, 최종 레이어 부근에서 한번 더 강력한 확률 변동이 나타난다. 

% (7) 확률의 강한 변동이 발생한 레이어 부근은 환각 탐지를 결정할 수 있는 정보 추출이나 최종 의사 결정을 나타내는 것으로 여겨진다~\citep{cheninside,jiang2024large}. 그러나, Figure 1의 LLMs는 pre- and post-negated example에 대해서 marginal한 차이만을 보인다. 이는 부정형으로 인해서 입력으로 주어진 문맥이나 지식이 변형되었음에도, 모델이 환각 여부를 판단함에 있어서 부정형으로 인한 변화를 명확하게 인지하지 못함을 암시한다. 이러한 현상은 negation을 모델 latent representation 내에서 logical operator 역할보다 하나의 단어로 여기면서 발생하는 것과 연관있다 \citep{bhar2024strong}. 

\paragraph{Negation Amplifies Hallucination Bias} To further analyze the above results, we examine the probability shifts across each layer for cases where hallucination judgments were correct versus incorrect for pre- and post-negated examples. As shown in Figure \ref{fig:figure2}, Llama3 and Mistral present substantial differences in probability shifts between successful and unsuccessful hallucination detection for pre-negated examples. However, for post-negated examples, these differences are relatively minor. The models exhibit stronger confidence in incorrect predictions compared to correct ones. These results demonstrate that when negated text is provided as input, models experience confusion in hallucination detection, highlighting the risk of falling into new hallucination patterns induced by negation.

% (8) 더욱이 post-negated example는 전반적으로 그들의 결정에 더 높은 확신을 가지거나 최종 레이어 부근에서 크게 변동한다. 이는 \citet{jiang2024large}의 실험 결과에서 나타난 환각을 발생하는 경우의 토큰 확률 변화와 유사하며, 부정형 예시에 대해서 환각으로 편향된 판단을 하는 원인으로 여겨짐.

% (9) 위의 결과에 대한 더 구체적인 분석을 위해 우리는 pre- and post-negated example에 대해서 환각 여부를 맞추는 경우와 틀린 경우의 the probability shift across each layer를 관찰한다.

% (10) As described in Figure 2, Llama 3와 Mistral은 pre-negated example인 경우 환각 탐지에 성공하는 경우와 그렇지 않은 경우의 확률 변동의 차이가 상당히 큼. 그러나, post-negated example인 경우에는 그 차이가 상대적으로 미미하게 나타남. 더욱이 오답에 대한 확신이 정답에 비해 더 강하게 나타난다. 이러한 결과는 부정형 텍스트가 입력으로 주어지는 경우에 모델은 환각 판별에 있어서 강한 혼란을 느끼며, 부정어로 인한 새로운 환각에 빠질 수 있는 위험성을 나타낸다. 

% (4) Final layer where the model makes final decoding decisions based on the information gathered from previous layers에 다가갈 수록 생성할 토큰에 대한 확신이 점차 강해진다. 

\paragraph{A3. Limited Improvements but Reveal Underlying Challenges}

\paragraph{In-Context Learning Shows Inconsistent Gains Across Models and Tasks}
Table \ref{tab:table8} compares model performance across 0-, 2-, and 4-shot settings, where the number of examples provided in the input prompt varies. On average, the 2-shot setting yields the highest performance across comparable cases. However, the influence of in-context examples on performance varies depending on the dataset and model, indicating that an increase in the number of examples does not necessarily guarantee improved performance. Interestingly, Llama3 demonstrates significant difficulty in following instructions to generate responses for hallucination detection in the zero-shot setting, resulting in notably low scores. This outcome shows that, as the number of shots increases, instruction-following abilities—beyond hallucination detection—also play a role in determining performance. Mistral exhibits relatively stable improvements across shots, though post-negated cases still reduce performance. Qwen3-4B shows stable instruction-following across shot settings, with strong pre-negated performance on BamBoo and SelfCheckGPT. However, it exhibits sharp declines under post-negated conditions, indicating particular sensitivity to negation despite otherwise consistent few-shot gains. These results imply that In-Context Learning can immediately enhance hallucination detection performance for certain models, datasets, and tasks, particularly when balanced examples with pre-/post-negated text and hallucination/no-hallucination cases are included. However, even in the 4-shot with balanced examples, performance declines were observed for hallucination detection with negated text. This raises questions about whether models fundamentally understand negated text and whether they can mitigate newly induced hallucinations.

% Table \ref{tab:table8} 은 입력 프롬프트 내에 제공된 예시의 갯수를 0, 2, 4-shot으로 설정하여 성능을 비교한다. 비교 가능한 케이스 중에서 2-shot이 평균적으로 높은 성능을 보인다. 그러나, 각 데이터와 모델에 따라서 context로 주어진 예시의 갯수가 성능에 영향이 다르며, increase in number of examples는 성능의 향상을 반드시 보장하지 않는다. 

% 흥미롭게도 Llama 3는 zero-shot에서 환각 판별을 위한 답변을 생성하는 instruction을 following하는데 어려움을 겪으면서 상당히 낮은 점수를 보인다. 이러한 결과는 실험에 사용한 모델들이 shot이 증가함에 따라서 환각 판별 이외의 instruction following의 요소도 성능에 영향을 미친다는 것을 보여준다. 

% 이러한 결과는 In-Context Learning은 특정한 모델과 데이터, 그리고 태스크에 대해서는 즉각적으로 환각 탐지 성능을 개선시킬 수 있으며, balanced examples with pre-/post-negated text and hallucination/no-hallucination cases가 경우에 따라서 효과를 발휘할 수 있음.

% 그러나, balanced examples를 지닌 4 shot setting에서도 부정형 텍스트에 대한 환각 탐지에서 일부 성능이 하락했으며, 본질적으로 부정형 텍스트에 대한 모델의 이해와 새로운 환각을 완화할 수 있을지는 의문이다.

\paragraph{CoT Reasoning Fails to Provide Consistent Improvements}
To address LLMs’ low hallucination detection capabilities when processing negated text, we applied CoT reasoning with in-context learning examples to enhance performance. Table \ref{tab:table9} presents the experimental results of applying CoT reasoning steps to 2-shot and 4-shot in-context learning settings. As described in $\S$\ref{sec:appendix_cot}, the models are instructed to articulate the reasoning behind their judgment during hallucination detection. The results demonstrate that the effectiveness of CoT reasoning varies significantly depending on the dataset, task, number of context examples, and the model used. Mistral and Llama2 show substantial performance improvement on pre-negated examples in HaluEval when using CoT prompts. However, consistent performance gains are not observed, with post-negated examples often showing minimal improvement or even performance degradation. These findings align with \citet{li2023halueval}, who argue that CoT reasoning steps alone are insufficient as a fundamental solution for improving hallucination detection.

% To address LLMs’ low hallucination detection capabilities in 부정형 텍스트가 주어졌을 경우, we apply CoT reasoning with in-context learning examples to improve performance. Table \ref{tab:table9}은 CoT reasoning step을 In-Context Learning에서 사용한 2-shot과 4-shot setting에 적용한 실험 결과를 나타낸다. As described in $\S$\ref{sec:analysis}, 우리는 환각 탐지 과정에서 judgement에 대한 이유를 서술하도록 하고, 판별을 하도록 함. The results show that CoT도 데이터, 태스크, context examples의 갯수 그리고 모델에 따라서 그 효과에 큰 차이. Mistral와 Llama2는 HaluEval에서 pre-negated에 대해서 CoT prompt로 큰 성능 개선이 나타나기도 함. 그러나, 일관된 성능 개선이 나타나지 않으며, post-negated examples에 대해서는 성능 개선이 크지 않으며 오히려 하락이 나타나기도 함. 이러한 결과는  
% \citet{li2023halueval}의 CoT reasoning steps are not sufficient as fundamental solutions for improving hallucination detection과 align한다. 
\begin{figure}[t]
\centering
\includegraphics[width=1\linewidth]{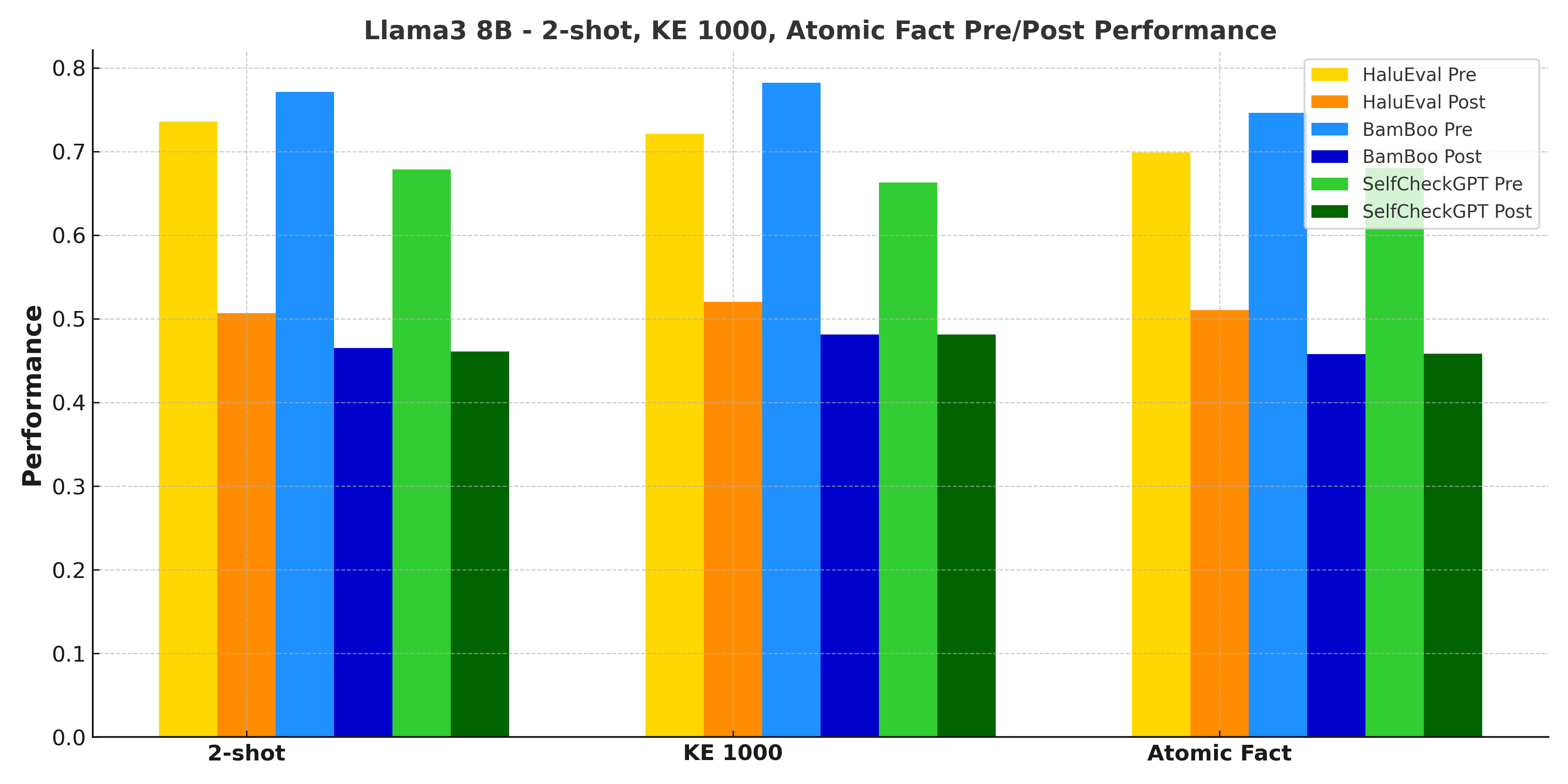}
\caption{Performance comparison of Llama3 across NegHalu subsets under negated knowledge updates using AlphaEdit and two different target corpus.}
 \label{fig:figure3}
\end{figure}

\paragraph{Knowledge Editing Modifies Model Behavior but Does Not Resolve the Negation Problem}
Based on the earlier empirical results, where negated text is misclassified as containing hallucinations, we apply knowledge editing to mitigate newly induced hallucinations or biases. To address this, we use negated knowledge—transformed from positively framed knowledge—to ensure consistency with real-world facts or commonsense. We employ AlphaEdit, a null-space-constrained knowledge editing, to update negated knowledge while minimizing damage to the model’s existing knowledge. Our experiments use two different editing target corpora: 1,000 factual statements from ROME’s dataset \citep{meng2022locating} (KE 1000) and atomic facts parsed from each dataset's given knowledge, transformed into negated knowledge (Atomic Fact). Figure \ref{fig:figure3} compares the performance of models in a 2-shot setting without knowledge editing versus with knowledge updated using AlphaEdit applied to KE 1000 or Atomic Fact corpus. The results show that knowledge editing with KE 1000 effectively minimizes damage to pre-negated knowledge while slightly improving performance on post-negated examples. However, this approach does not fundamentally resolve issues caused by negated text, as performance degradation on pre-negated examples appears in SelfCheckGPT, showing limitations in handling negated knowledge effectively.

% 우리는 부정형 텍스트가 환각 요소를 포함하는 것으로 판단하는 앞선 empirical results를 바탕으로, 새로운 환각 또는 bias를 완화하기 위해서 지식 편집 방식을 적용한다. 우리는 기존에 긍정형으로 표현된 지식이 실제 세계의 사실이나 commonsense를 위배하지 않도록 의미를 변형한 Negated knowledge를 활용한다. 이러한 목적을 달성하기 위해 우리는 null-space constrained 지식 편집 방식인 AlphaEdit을 활용하여, 모델의 기존 지식에 대한 손상을 최소화하고 부정형 지식을 업데이트 한다. 
% 우리는 ROME에서 활용한 1000개의 factual statements \citep{meng2022locating} (KE 1000)과 각 데이터셋의 given knowledge를 atomic fact로 파싱하고 부정형 지식으로 변형 (Atomic Fact)하여 각기 다른 편집 타겟 코퍼스로 사용한다. Figure \ref{fig:figure3}은 지식 편집 없이 2-shot인 경우의 성능과 KE 1000 또는 Atomic Fact를 편집 코퍼스로 활용하여 부정형 지식을 AlphaEdit으로 업데이트 하는 경우의 성능을 비교한다. 실험 결과는 KE 1000을 활용한 지식 편집이 pre-negated 지식에 대한 손상을 최소화하며 post-negated에 대한 성능을 소폭 개선할 수 있음을 나타낸다. 그러나, 본질적으로 부정형 텍스트로 인한 문제를 해결할 수 있는 방법으로 보기는 어려우며, SelfCheckGPT에서는 pre-negated에서 성능이 감소하는 문제가 발생함. 

\begin{table}[h]
\centering
\resizebox{1.0\columnwidth}{!}{
\renewcommand{\arraystretch}{1.2}
\begin{tabular}{lccc}
\toprule
\multirow{2}{*}{\textbf{NegHalu+}} & \multicolumn{3}{c}{\textbf{HaluEval (Pre/Post) - Accuracy}} \\
\cmidrule(lr){2-4}
& \textbf{QA} & \textbf{Dialogue} & \textbf{Summarization} \\
\midrule
Llama-2-7B & 0.4825 / 0.4125 & 0.6150 / 0.5650 & 0.4750 / 0.5250 \\
Llama-3-8B & 0.7650 / 0.4525 & 0.7825 / 0.5150 &  0.6600 / 0.5675 \\
Mistral-7B-v0.3 & 0.5900 / 0.5075 & 0.7050 / 0.5575 & 0.5950 / 0.5025 \\
Qwen3-4B & 0.4425 / 0.2775 & 0.7425 / 0.5150 & 0.5625 / 0.4675 \\
\bottomrule
\end{tabular}}
\caption{Performance of models on HaluEval where originally explicit negation examples were partially replaced (5\% each) with implicit and morphological negation.}
\label{tab:neghalu_plus}
\end{table}
\section{Effect of Adding Implicit and Morphological Negation}
Table~\ref{tab:neghalu_plus} presents the results after extending the originally explicit-only HaluEval with a small proportion of implicit and morphological negation. 
Implicit forms include expressions such as \textit{doubt}, \textit{hardly}, \textit{fails to}, \textit{unlikely that}, and \textit{questionable whether}. 
Morphological forms use affixes such as \textit{un}, \textit{in}, \textit{im}, and \textit{dis}, as well as the suffix \textit{less}, to produce words like \textit{incorrect}, \textit{impossible}, and \textit{useless}.  

The overall pattern remains clear. Post-negated inputs reduce accuracy across models and tasks, but the scale and distribution of the decline shift once diverse negation types are introduced. Llama2 shows a smaller gap between pre- and post-negated inputs, with summarization even improving from 47.5\% to 52.5\%, which suggests partial robustness when negation cues vary. Llama3 continues to display the strongest sensitivity, while Mistral shows slightly milder degradation on dialogue. Qwen3-4B still suffers sharp declines in QA and dialogue, revealing a consistent vulnerability to negation regardless of its form.

NegHalu+ maintains the overall effect of negation but reveals distinct failure profiles compared with the explicit-only setting. Some models such as Llama2 and Mistral show slightly reduced performance gaps, while others such as Llama3 and Qwen3-4B exhibit sharper declines, especially on QA and dialogue. These results indicate that robustness measured only with explicit negation underestimates true vulnerability, and that a mixture of negation types exposes model-specific weaknesses across tasks.

\section{Conclusion}
In this study, we explore the impact of negated text on hallucination detection in LLMs by constructing NegHalu, a dataset designed to evaluate model performance under pre- and post-negation scenarios. Through systematic experiments, we examine three key research questions and uncover fundamental limitations in how LLMs process negation. Our answers highlight that LLMs exhibit performance degradation, systematic biases, internal behavioral constraints, and limited performance improvements in handling hallucinations. Furthermore, we identify the risk of new types of hallucinations emerging due to negation, posing additional challenges for model reliability. For future work, we emphasize the need for deeper architectural refinements and advanced strategies to improve LLMs' ability to process negation effectively, ensuring robust and reliable performance across diverse contexts.

\section*{Limitations}
This study has several limitations that warrant consideration. First, we use only a subset of the HaluEval dataset, which may lead to results for pre-negated text differing slightly from those obtained using the full dataset. Additionally, the BamBoo dataset contains an insufficient number of samples to achieve balanced labels, resulting in experiments being conducted with slight label imbalance. Second, due to computational resource constraints, we were unable to compare larger models. Even if such experiments were conducted, significant differences in hyperparameter settings would be required, which could lead to outcomes different from those reported here. Third, during the creation and verification of NegHalu, there may be a small number of errors in labeling hallucinations that differ from human interpretations. However, these are not considered substantial enough to overturn the overall experimental findings. Fourth, while our verification step uses multiple GPT-4 models, we acknowledge that employing the same model as both generator and judge could introduce bias in favor of its own generations. To mitigate this risk, we ensembled models with different temperature settings and required full agreement among them. In addition, the authors manually reviewed and adjusted the outputs where necessary. Fifth, while we explored various methods to address the hallucinations and biases introduced by the negated text, we were unable to propose a complete solution. We consider solutions targeting only the impact of negated text on hallucination problems to be impractical. Instead of relying on external modules or extensive tuning, we apply intrinsic knowledge and minimal knowledge editing. We hope that the analyses presented in this paper will provide a solid foundation for future research. Lastly, we excluded ROME \citep{meng2022locating} and MEMIT \citep{meng2023massediting} from our experiments, as their application resulted in significant performance degradation on pre-negated knowledge, unlike AlphaEdit.

% This study has several limitations that should be considered. First, we utilize HaluEval dataset의 전체가 아닌 일부를 사용하여, pre-negated text에 대한 실험 결과가 전체를 사용하는 경우와 다소 차이가 발생할 수 있음. 또한, BamBoo의 데이터셋은 레이블 밸런스를 맞추기에는 수량이 너무 적어서, 약간 불균형한 상태로 실험을 진행함. Second, 컴퓨팅 리소스의 제약으로 더 큰 모델에 대한 비교를 하기 어려웠으며, 만약 실험을 한다고 하여도 하이퍼 파라미터 세팅에서 큰 차이를 내야하므로 포함시키지 않았음. 따라서, 더 큰 모델의 경우 우리가 레포트하는 실험 결과와 다른 양상이 발생할 수 있음. 세 번째, NegHalu를 제작하고 Curation 하는 과정에서 사람이 생각하는 환각과는 다소 차이가 존재하는 소량의 오류가 포함되어 있을 수 있음. 다만, 이는 전체적인 실험 결과를 뒤집을 정도의 차이로 여겨지지 않음. 네 번째, 우리는 다양한 방법론을 바탕으로 부정형 텍스트로 인한 새로운 환각 또는 편향 문제를 해결하고자 하였으나 완전한 솔루션을 제시하지 못함. 우리는 이 논문의 분석 결과가 이후에 있을 연구에 중요한 기틀이 되길 희망함. 마지막으로 AlphaEdit 이외의 ROME과 MEMIT은 pre-negated knowledge에 대한 심각한 성능 손상이 발생하여 포함시키지 않았음. 

\section*{Acknowledgments}
This work was supported by Institute for Information \& communications Technology Promotion(IITP) grant funded by the Korea government(MSIT) 
(RS-2024-00398115, Research on the reliability and coherence of outcomes produced by Generative AI). This research was supported by Basic Science Research Program through the National Research Foundation of Korea(NRF) funded by the Ministry of Education(NRF-2021R1A6A1A03045425). This work was supported by the Commercialization Promotion Agency for R\&D Outcomes(COMPA) grant funded by the Korea government(Ministry of Science and ICT)(2710086166). 

\bibliography{custom}

\begin{thebibliography}{50}
\providecommand{\natexlab}[1]{#1}

\bibitem[{AI@Meta(2024)}]{llama3modelcard}
AI@Meta. 2024.
\newblock \href {https://github.com/meta-llama/llama3/blob/main/MODEL_CARD.md} {Llama 3 model card}.

\bibitem[{Ainslie et~al.(2023)Ainslie, Lee-Thorp, de~Jong, Zemlyanskiy, Lebron, and Sanghai}]{ainslie2023gqa}
Joshua Ainslie, James Lee-Thorp, Michiel de~Jong, Yury Zemlyanskiy, Federico Lebron, and Sumit Sanghai. 2023.
\newblock Gqa: Training generalized multi-query transformer models from multi-head checkpoints.
\newblock In \emph{Proceedings of the 2023 Conference on Empirical Methods in Natural Language Processing}, pages 4895--4901.

\bibitem[{Arnaout et~al.(2022)Arnaout, Razniewski, Weikum, and Pan}]{arnaout2022uncommonsense}
Hiba Arnaout, Simon Razniewski, Gerhard Weikum, and Jeff~Z Pan. 2022.
\newblock Uncommonsense: Informative negative knowledge about everyday concepts.
\newblock In \emph{Proceedings of the 31st ACM International Conference on Information \& Knowledge Management}, pages 37--46.

\bibitem[{Bai et~al.(2023)Bai, Bai, Chu, Cui, Dang, Deng, Fan, Ge, Han, Huang et~al.}]{bai2023qwen}
Jinze Bai, Shuai Bai, Yunfei Chu, Zeyu Cui, Kai Dang, Xiaodong Deng, Yang Fan, Wenbin Ge, Yu~Han, Fei Huang, et~al. 2023.
\newblock Qwen technical report.
\newblock \emph{arXiv preprint arXiv:2309.16609}.

\bibitem[{Beltagy et~al.(2020)Beltagy, Peters, and Cohan}]{beltagy2020longformer}
Iz~Beltagy, Matthew~E Peters, and Arman Cohan. 2020.
\newblock Longformer: The long-document transformer.
\newblock \emph{arXiv preprint arXiv:2004.05150}.

\bibitem[{Bhar and Asher(2024)}]{bhar2024strong}
Swarnadeep Bhar and Nicholas Asher. 2024.
\newblock Strong hallucinations from negation and how to fix them.
\newblock In \emph{Findings of the Association for Computational Linguistics ACL 2024}, pages 12670--12687.

\bibitem[{Brown et~al.(2020)Brown, Mann, Ryder, Subbiah, Kaplan, Dhariwal, Neelakantan, Shyam, Sastry, Askell et~al.}]{brown2020language}
Tom Brown, Benjamin Mann, Nick Ryder, Melanie Subbiah, Jared~D Kaplan, Prafulla Dhariwal, Arvind Neelakantan, Pranav Shyam, Girish Sastry, Amanda Askell, et~al. 2020.
\newblock Language models are few-shot learners.
\newblock \emph{Advances in neural information processing systems}, 33:1877--1901.

\bibitem[{Chen et~al.(2024)Chen, Liu, Chen, Gu, Wu, Tao, Fu, and Ye}]{cheninside}
Chao Chen, Kai Liu, Ze~Chen, Yi~Gu, Yue Wu, Mingyuan Tao, Zhihang Fu, and Jieping Ye. 2024.
\newblock \href {https://openreview.net/forum?id=Zj12nzlQbz} {{INSIDE}: {LLM}s' internal states retain the power of hallucination detection}.
\newblock In \emph{The Twelfth International Conference on Learning Representations}.

\bibitem[{Chen et~al.(2023)Chen, Shi, Fu, Cheng, Li, and Xiao}]{chen2023say}
Jiangjie Chen, Wei Shi, Ziquan Fu, Sijie Cheng, Lei Li, and Yanghua Xiao. 2023.
\newblock Say what you mean! large language models speak too positively about negative commonsense knowledge.
\newblock In \emph{Proceedings of the 61st Annual Meeting of the Association for Computational Linguistics (Volume 1: Long Papers)}, pages 9890--9908.

\bibitem[{Choubey et~al.(2023)Choubey, Fabbri, Vig, Wu, Liu, and Rajani}]{choubey2023cape}
Prafulla~Kumar Choubey, Alex Fabbri, Jesse Vig, Chien-Sheng Wu, Wenhao Liu, and Nazneen Rajani. 2023.
\newblock Cape: Contrastive parameter ensembling for reducing hallucination in abstractive summarization.
\newblock In \emph{Findings of the Association for Computational Linguistics: ACL 2023}, pages 10755--10773.

\bibitem[{Chuang et~al.(2023)Chuang, Xie, Luo, Kim, Glass, and He}]{chuang2023dola}
Yung-Sung Chuang, Yujia Xie, Hongyin Luo, Yoon Kim, James~R Glass, and Pengcheng He. 2023.
\newblock Dola: Decoding by contrasting layers improves factuality in large language models.
\newblock In \emph{The Twelfth International Conference on Learning Representations}.

\bibitem[{Dale et~al.(2023)Dale, Voita, Barrault, and Costa-juss{\`a}}]{dale2023detecting}
David Dale, Elena Voita, Lo{\"\i}c Barrault, and Marta~R Costa-juss{\`a}. 2023.
\newblock Detecting and mitigating hallucinations in machine translation: Model internal workings alone do well, sentence similarity even better.
\newblock In \emph{Proceedings of the 61st Annual Meeting of the Association for Computational Linguistics (Volume 1: Long Papers)}, pages 36--50.

\bibitem[{Dar et~al.(2023)Dar, Geva, Gupta, and Berant}]{dar2023analyzing}
Guy Dar, Mor Geva, Ankit Gupta, and Jonathan Berant. 2023.
\newblock Analyzing transformers in embedding space.
\newblock In \emph{Proceedings of the 61st Annual Meeting of the Association for Computational Linguistics (Volume 1: Long Papers)}, pages 16124--16170.

\bibitem[{Dong et~al.(2024{\natexlab{a}})Dong, Li, Dai, Zheng, Ma, Li, Xia, Xu, Wu, Chang et~al.}]{dong2024survey}
Qingxiu Dong, Lei Li, Damai Dai, Ce~Zheng, Jingyuan Ma, Rui Li, Heming Xia, Jingjing Xu, Zhiyong Wu, Baobao Chang, et~al. 2024{\natexlab{a}}.
\newblock A survey on in-context learning.
\newblock In \emph{Proceedings of the 2024 Conference on Empirical Methods in Natural Language Processing}, pages 1107--1128.

\bibitem[{Dong et~al.(2024{\natexlab{b}})Dong, Tang, Li, Zhao, and Wen}]{dong2024bamboo}
Zican Dong, Tianyi Tang, Junyi Li, Wayne~Xin Zhao, and Ji-Rong Wen. 2024{\natexlab{b}}.
\newblock Bamboo: A comprehensive benchmark for evaluating long text modeling capacities of large language models.
\newblock In \emph{Proceedings of the 2024 Joint International Conference on Computational Linguistics, Language Resources and Evaluation (LREC-COLING 2024)}, pages 2086--2099.

\bibitem[{Fang et~al.(2024)Fang, Jiang, Wang, Ma, Wang, He, and Chua}]{fang2024alphaedit}
Junfeng Fang, Houcheng Jiang, Kun Wang, Yunshan Ma, Xiang Wang, Xiangnan He, and Tat-seng Chua. 2024.
\newblock Alphaedit: Null-space constrained knowledge editing for language models.
\newblock \emph{arXiv preprint arXiv:2410.02355}.

\bibitem[{Farquhar et~al.(2024)Farquhar, Kossen, Kuhn, and Gal}]{farquhar2024detecting}
Sebastian Farquhar, Jannik Kossen, Lorenz Kuhn, and Yarin Gal. 2024.
\newblock Detecting hallucinations in large language models using semantic entropy.
\newblock \emph{Nature}, 630(8017):625--630.

\bibitem[{Gubelmann and Handschuh(2022)}]{gubelmann2022context}
Reto Gubelmann and Siegfried Handschuh. 2022.
\newblock Context matters: A pragmatic study of plms’ negation understanding.
\newblock In \emph{Proceedings of the 60th Annual Meeting of the Association for Computational Linguistics (Volume 1: Long Papers)}, pages 4602--4621.

\bibitem[{Guerreiro et~al.(2023)Guerreiro, Alves, Waldendorf, Haddow, Birch, Colombo, and Martins}]{guerreiro-etal-2023-hallucinations}
Nuno~M. Guerreiro, Duarte~M. Alves, Jonas Waldendorf, Barry Haddow, Alexandra Birch, Pierre Colombo, and Andr{\'e} F.~T. Martins. 2023.
\newblock \href {https://doi.org/10.1162/tacl_a_00615} {Hallucinations in large multilingual translation models}.
\newblock \emph{Transactions of the Association for Computational Linguistics}, 11:1500--1517.

\bibitem[{Hossain et~al.(2022)Hossain, Chinnappa, and Blanco}]{hossain2022analysis}
Md~Mosharaf Hossain, Dhivya Chinnappa, and Eduardo Blanco. 2022.
\newblock An analysis of negation in natural language understanding corpora.
\newblock In \emph{Proceedings of the 60th Annual Meeting of the Association for Computational Linguistics (Volume 2: Short Papers)}, pages 716--723.

\bibitem[{Huang et~al.(2023{\natexlab{a}})Huang, Yu, Ma, Zhong, Feng, Wang, Chen, Peng, Feng, Qin et~al.}]{huang2023survey}
Lei Huang, Weijiang Yu, Weitao Ma, Weihong Zhong, Zhangyin Feng, Haotian Wang, Qianglong Chen, Weihua Peng, Xiaocheng Feng, Bing Qin, et~al. 2023{\natexlab{a}}.
\newblock A survey on hallucination in large language models: Principles, taxonomy, challenges, and open questions.
\newblock \emph{arXiv preprint arXiv:2311.05232}.

\bibitem[{Huang et~al.(2023{\natexlab{b}})Huang, Song, Wang, Zhao, Chen, Juefei-Xu, and Ma}]{huang2023look}
Yuheng Huang, Jiayang Song, Zhijie Wang, Shengming Zhao, Huaming Chen, Felix Juefei-Xu, and Lei Ma. 2023{\natexlab{b}}.
\newblock Look before you leap: An exploratory study of uncertainty measurement for large language models.
\newblock \emph{arXiv preprint arXiv:2307.10236}.

\bibitem[{Ji et~al.(2023{\natexlab{a}})Ji, Lee, Frieske, Yu, Su, Xu, Ishii, Bang, Madotto, and Fung}]{ji2023survey}
Ziwei Ji, Nayeon Lee, Rita Frieske, Tiezheng Yu, Dan Su, Yan Xu, Etsuko Ishii, Ye~Jin Bang, Andrea Madotto, and Pascale Fung. 2023{\natexlab{a}}.
\newblock Survey of hallucination in natural language generation.
\newblock \emph{ACM Computing Surveys}, 55(12):1--38.

\bibitem[{Ji et~al.(2023{\natexlab{b}})Ji, Liu, Lee, Yu, Wilie, Zeng, and Fung}]{ji2023rho}
Ziwei Ji, Zihan Liu, Nayeon Lee, Tiezheng Yu, Bryan Wilie, Min Zeng, and Pascale Fung. 2023{\natexlab{b}}.
\newblock Rho: Reducing hallucination in open-domain dialogues with knowledge grounding.
\newblock In \emph{Findings of the Association for Computational Linguistics: ACL 2023}, pages 4504--4522.

\bibitem[{Jiang et~al.(2023)Jiang, Sablayrolles, Mensch, Bamford, Chaplot, Casas, Bressand, Lengyel, Lample, Saulnier et~al.}]{jiang2023mistral}
Albert~Q Jiang, Alexandre Sablayrolles, Arthur Mensch, Chris Bamford, Devendra~Singh Chaplot, Diego de~las Casas, Florian Bressand, Gianna Lengyel, Guillaume Lample, Lucile Saulnier, et~al. 2023.
\newblock \href {https://mistral.ai/news/announcing-mistral-7b/} {Mistral 7b}.
\newblock \emph{arXiv preprint arXiv:2310.06825}.

\bibitem[{Jiang et~al.(2024)Jiang, Qi, Hong, Fu, Cheng, Meng, Yu, Zhou, and Zhou}]{jiang2024large}
Che Jiang, Biqing Qi, Xiangyu Hong, Dayuan Fu, Yang Cheng, Fandong Meng, Mo~Yu, Bowen Zhou, and Jie Zhou. 2024.
\newblock On large language models’ hallucination with regard to known facts.
\newblock In \emph{Proceedings of the 2024 Conference of the North American Chapter of the Association for Computational Linguistics: Human Language Technologies (Volume 1: Long Papers)}, pages 1041--1053.

\bibitem[{Kassner and Sch{\"u}tze(2020)}]{kassner2020negated}
Nora Kassner and Hinrich Sch{\"u}tze. 2020.
\newblock Negated and misprimed probes for pretrained language models: Birds can talk, but cannot fly.
\newblock In \emph{Proceedings of the 58th Annual Meeting of the Association for Computational Linguistics}, pages 7811--7818.

\bibitem[{Khandelwal and Sawant(2020)}]{khandelwal2020negbert}
Aditya Khandelwal and Suraj Sawant. 2020.
\newblock Negbert: A transfer learning approach for negation detection and scope resolution.
\newblock In \emph{Proceedings of the Twelfth Language Resources and Evaluation Conference}, pages 5739--5748.

\bibitem[{Li et~al.(2023)Li, Cheng, Zhao, Nie, and Wen}]{li2023halueval}
Junyi Li, Xiaoxue Cheng, Wayne~Xin Zhao, Jian-Yun Nie, and Ji-Rong Wen. 2023.
\newblock Halueval: A large-scale hallucination evaluation benchmark for large language models.
\newblock In \emph{Proceedings of the 2023 Conference on Empirical Methods in Natural Language Processing}, pages 6449--6464.

\bibitem[{Magesh et~al.(2024)Magesh, Surani, Dahl, Suzgun, Manning, and Ho}]{magesh2024hallucination}
Varun Magesh, Faiz Surani, Matthew Dahl, Mirac Suzgun, Christopher~D Manning, and Daniel~E Ho. 2024.
\newblock Hallucination-free? assessing the reliability of leading ai legal research tools.
\newblock \emph{arXiv preprint arXiv:2405.20362}.

\bibitem[{Manakul et~al.(2023)Manakul, Liusie, and Gales}]{manakul2023selfcheckgpt}
Potsawee Manakul, Adian Liusie, and Mark Gales. 2023.
\newblock Selfcheckgpt: Zero-resource black-box hallucination detection for generative large language models.
\newblock In \emph{Proceedings of the 2023 Conference on Empirical Methods in Natural Language Processing}, pages 9004--9017.

\bibitem[{Meng et~al.(2022)Meng, Bau, Andonian, and Belinkov}]{meng2022locating}
Kevin Meng, David Bau, Alex Andonian, and Yonatan Belinkov. 2022.
\newblock Locating and editing factual associations in gpt.
\newblock \emph{Advances in Neural Information Processing Systems}, 35:17359--17372.

\bibitem[{Meng et~al.(2023)Meng, Sharma, Andonian, Belinkov, and Bau}]{meng2023massediting}
Kevin Meng, Arnab~Sen Sharma, Alex~J Andonian, Yonatan Belinkov, and David Bau. 2023.
\newblock \href {https://openreview.net/forum?id=MkbcAHIYgyS} {Mass-editing memory in a transformer}.
\newblock In \emph{The Eleventh International Conference on Learning Representations}.

\bibitem[{Minsky(1997)}]{minsky1997negative}
Marvin Minsky. 1997.
\newblock Negative expertise.
\newblock In \emph{Expertise in context: human and machine}, pages 515--521.

\bibitem[{Morante et~al.(2011)Morante, Schrauwen, and Daelemans}]{morante2011annotation}
Roser Morante, Sarah Schrauwen, and Walter Daelemans. 2011.
\newblock Annotation of negation cues and their scope: Guidelines v1.
\newblock \emph{Computational linguistics and psycholinguistics technical report series, CTRS-003}, pages 1--42.

\bibitem[{nostalgebraist(2020)}]{nostalgebraist2020}
nostalgebraist. 2020.
\newblock \href {https://www.lesswrong.com/posts/AcKRB8wDpdaN6v6ru/interpreting-gpt-the-logit-lens} {Interpreting gpt: the logit lens}.

\bibitem[{OpenAI(2023)}]{openai2023gpt4}
OpenAI. 2023.
\newblock \href {https://arxiv.org/abs/2303.08774} {Gpt-4 technical report}.
\newblock \emph{Preprint}, arXiv:2303.08774.

\bibitem[{Seo et~al.(2024)Seo, Lee, Park, Hong, Lee, and Lim}]{seo2024kocommongen}
Jaehyung Seo, Jaewook Lee, Chanjun Park, SeongTae Hong, Seungjun Lee, and Heui-Seok Lim. 2024.
\newblock Kocommongen v2: A benchmark for navigating korean commonsense reasoning challenges in large language models.
\newblock In \emph{Findings of the Association for Computational Linguistics ACL 2024}, pages 2390--2415.

\bibitem[{Su et~al.(2024)Su, Ahmed, Lu, Pan, Bo, and Liu}]{su2024roformer}
Jianlin Su, Murtadha Ahmed, Yu~Lu, Shengfeng Pan, Wen Bo, and Yunfeng Liu. 2024.
\newblock Roformer: Enhanced transformer with rotary position embedding.
\newblock \emph{Neurocomputing}, 568:127063.

\bibitem[{Team et~al.(2024)Team, Mesnard, Hardin, Dadashi, Bhupatiraju, Pathak, Sifre, Rivi{\`e}re, Kale, Love et~al.}]{team2024gemma}
Gemma Team, Thomas Mesnard, Cassidy Hardin, Robert Dadashi, Surya Bhupatiraju, Shreya Pathak, Laurent Sifre, Morgane Rivi{\`e}re, Mihir~Sanjay Kale, Juliette Love, et~al. 2024.
\newblock \href {https://doi.org/10.48550/arXiv.2403.08295} {Gemma: Open models based on gemini research and technology}.
\newblock \emph{arXiv preprint arXiv:2403.08295}.

\bibitem[{Touvron et~al.(2023)Touvron, Martin, Stone, Albert, Almahairi, Babaei, Bashlykov, Batra, Bhargava, Bhosale et~al.}]{touvron2023llama}
Hugo Touvron, Louis Martin, Kevin Stone, Peter Albert, Amjad Almahairi, Yasmine Babaei, Nikolay Bashlykov, Soumya Batra, Prajjwal Bhargava, Shruti Bhosale, et~al. 2023.
\newblock \href {https://doi.org/10.48550/arXiv.2307.09288} {Llama 2: Open foundation and fine-tuned chat models}.
\newblock \emph{arXiv preprint arXiv:2307.09288}.

\bibitem[{Truong et~al.(2023)Truong, Baldwin, Verspoor, and Cohn}]{truong2023language}
Thinh~Hung Truong, Timothy Baldwin, Karin Verspoor, and Trevor Cohn. 2023.
\newblock Language models are not naysayers: an analysis of language models on negation benchmarks.
\newblock In \emph{Proceedings of the 12th Joint Conference on Lexical and Computational Semantics (* SEM 2023)}, pages 101--114.

\bibitem[{van Son et~al.(2016)van Son, Van~Miltenburg, and Morante}]{van2016building}
Chantal van Son, Emiel Van~Miltenburg, and Roser Morante. 2016.
\newblock Building a dictionary of affixal negations.
\newblock In \emph{Proceedings of the Workshop on Extra-Propositional Aspects of Meaning in Computational Linguistics (ExProM)}, pages 49--56.

\bibitem[{Vanek et~al.(2024)Vanek, Mati{\'c}~{\v{S}}kori{\'c}, Ko{\v{s}}utar, Mat{\v{e}}jka, and Stone}]{vanek2024mental}
Norbert Vanek, Ana Mati{\'c}~{\v{S}}kori{\'c}, Sara Ko{\v{s}}utar, {\v{S}}t{\v{e}}p{\'a}n Mat{\v{e}}jka, and Kate Stone. 2024.
\newblock Mental simulation of the factual and the illusory in negation processing: evidence from anticipatory eye movements on a blank screen.
\newblock \emph{Scientific reports}, 14(1):2844.

\bibitem[{Wei et~al.(2022)Wei, Wang, Schuurmans, Bosma, Xia, Chi, Le, Zhou et~al.}]{wei2022chain}
Jason Wei, Xuezhi Wang, Dale Schuurmans, Maarten Bosma, Fei Xia, Ed~Chi, Quoc~V Le, Denny Zhou, et~al. 2022.
\newblock Chain-of-thought prompting elicits reasoning in large language models.
\newblock \emph{Advances in neural information processing systems}, 35:24824--24837.

\bibitem[{Yang et~al.(2023)Yang, Sun, and Wan}]{yang2023new}
Shiping Yang, Renliang Sun, and Xiaojun Wan. 2023.
\newblock A new benchmark and reverse validation method for passage-level hallucination detection.
\newblock In \emph{Findings of the Association for Computational Linguistics: EMNLP 2023}, pages 3898--3908.

\bibitem[{Ye et~al.(2023)Ye, Kuribayashi, Suzuki, Kobayashi, and Funayama}]{ye2023assessing}
Mengyu Ye, Tatsuki Kuribayashi, Jun Suzuki, Goro Kobayashi, and Hiroaki Funayama. 2023.
\newblock Assessing step-by-step reasoning against lexical negation: A case study on syllogism.
\newblock In \emph{Proceedings of the 2023 Conference on Empirical Methods in Natural Language Processing}, pages 14753--14773.

\bibitem[{Zhang and Sennrich(2019)}]{zhang2019root}
Biao Zhang and Rico Sennrich. 2019.
\newblock Root mean square layer normalization.
\newblock \emph{Advances in Neural Information Processing Systems}, 32.

\bibitem[{Zhang et~al.(2023)Zhang, Li, Cui, Cai, Liu, Fu, Huang, Zhao, Zhang, Chen et~al.}]{zhang2023siren}
Yue Zhang, Yafu Li, Leyang Cui, Deng Cai, Lemao Liu, Tingchen Fu, Xinting Huang, Enbo Zhao, Yu~Zhang, Yulong Chen, et~al. 2023.
\newblock Siren's song in the ai ocean: a survey on hallucination in large language models.
\newblock \emph{arXiv preprint arXiv:2309.01219}.

\bibitem[{Zhao et~al.(2020)Zhao, Cohen, and Webber}]{zhao2020reducing}
Zheng Zhao, Shay~B Cohen, and Bonnie Webber. 2020.
\newblock Reducing quantity hallucinations in abstractive summarization.
\newblock In \emph{Findings of the Association for Computational Linguistics: EMNLP 2020}, pages 2237--2249.

\end{thebibliography}

\appendix
\section{Summary of Research Questions and Answers}
\label{sec:appendix_rq}
We summarize the key findings of our study by revisiting the research questions and their corresponding answers.

\textbf{RQ1.} \textit{Can LLMs distinguish between hallucinations and faithful statements in negated text as effectively as in affirmative text?}

- \textbf{A1. LLMs exhibit performance degradation and bias in hallucination detection for negated text.} Our results show that LLMs struggle to maintain consistent hallucination detection performance in negated text, displaying a bias toward hallucination predictions in post-negated cases. Across all evaluated tasks—including QA, dialogue, and summarization—models incorrectly classify negated factual statements as hallucinations at significantly higher rates. This effect is observed regardless of model type, suggesting that negation disrupts LLMs’ hallucination judgments in a systematic and task-agnostic manner. 

\textbf{RQ2.} \textit{Can the model internally recognize differences caused by negation when detecting hallucinations?}

- \textbf{A2. LLMs show a lack of distinction between pre- and post-negated text.} Through logit lens analysis, we observe that LLMs exhibit strong probability shifts in the middle and final layers when processing negated inputs, yet these shifts do not result in meaningful internal differentiation between pre- and post-negated statements. Instead, negation appears to function as a lexical modifier rather than a logical transformation, leading to overgeneralized hallucination judgments. Additionally, models display increased confidence in incorrect predictions, reinforcing the idea that negation is not properly incorporated into the model’s reasoning process.
13
\textbf{RQ3.} \textit{Can targeted intervention strategies improve hallucination detection in the negated text?}

- \textbf{A3. Limited improvements but revealing underlying challenges.} We explore in-context learning, CoT reasoning, and knowledge editing as potential mitigation strategies. However, none of these methods provided a fundamental solution to negation-induced hallucination biases. In-context learning showed inconsistent performance gains, with improvements depending more on instruction-following ability rather than genuine negation comprehension. CoT reasoning improved pre-negated hallucination detection in certain cases but failed to generalize across datasets and did not consistently improve performance in post-negated cases. Knowledge editing slightly reduced hallucination errors in some conditions but failed to eliminate systematic negation biases, suggesting that negation errors are deeply embedded within the model’s internal representations rather than merely arising from incorrect factual knowledge.

\section{Experimental Details}
\label{sec:appendix}

To evaluate the LLMs, we used a single NVIDIA A6000 GPU with 48GB memory capacity and AMD EPYC 7513 32-core Processor CPUs.

\paragraph{Model Details}
For the NegHalu experiments, we followed the hyperparameter settings defined by the existing benchmark datasets. Across all datasets, we used greedy decoding without sampling methods. The maximum output length was set to 4096 for HaluEval, 32 for BamBoo, and 5 for SelfCheckGPT-WikiBio.

\textbf{Llama2} \citep{touvron2023llama} (\textit{meta-llama/Llama-2-7b-chat-hf}) is a Transformer-based language model with 7 billion parameters. This model employs the SwiGLU activation function, Rotary Position Embedding (RoPE) \citep{su2024roformer}, and RMSNorm \citep{zhang2019root} to enhance stability. Its configuration includes a maximum token length of 4096, 32 attention heads, 32 hidden layers, a vocabulary size of 32,000, and float16.

\textbf{Llama3} \citep{llama3modelcard} (\textit{meta-llama/Meta-Llama-3-8B-Instruct}) is a Transformer-based language model with 8 billion parameters, sharing the same fundamental structure as Llama2. It supports a maximum token length of 4096, 32 attention heads, 32 hidden layers, a vocabulary size of 128,256, and bfloat16.

\textbf{Mistral} \citep{jiang2023mistral} (\textit{mistralai/Mistral-7B-Instruct-v0.3}) is a Transformer-based language model with 7.3 billion parameters. It leverages Grouped-Query Attention (GQA) \citep{ainslie2023gqa} and Sliding Window Attention (SWA) \citep{beltagy2020longformer} mechanisms for computational efficiency. Its configuration includes a maximum token length of 4096, 32 attention heads, 32 hidden layers, a vocabulary size of 32,768, and bfloat16.

\textbf{Qwen3} \citep{bai2023qwen} (\textit{Qwen/Qwen3-4B}) is a Transformer-based language model with 4 billion parameters. It adopts RoPE \citep{su2024roformer}, GQA \citep{ainslie2023gqa}, and RMSNorm \citep{zhang2019root} to improve computational efficiency and stability. The model is configured with a maximum token length of 32,768 (extendable up to 131,072 with YaRN), 36 hidden layers, 32 attention heads with 8 key-value heads, and a vocabulary size of 151,936. We used the dense variant of Qwen3-4B with bfloat16 precision. In our NegHalu experiments, the model was run without enabling the \textit{think} mode, using only direct response generation under greedy decoding.

\paragraph{Knowledge Editing Details}
To update negated knowledge in Llama3 using AlphaEdit \citep{fang2024alphaedit}, we target layers 4 through 8 based on causal tracing results derived from 100,000 triplets obtained from Wikipedia \citep{meng2023massediting}. The editing corpus includes two sources: \textbf{KE 1000}, which consists of 1,000 factual statements from ROME \citep{meng2022locating}, and \textbf{Atomic Fact}, which represents up to 1,000 atomic facts parsed from the given knowledge in each dataset. To parse these facts and transform them into negated knowledge, we use the GPT-4 omni (\texttt{gpt-4o-2024-08-06}) \citep{openai2023gpt4} API. The prompts used for Atomic Fact parsing and negated knowledge construction are described in Tables \ref{tab:table10} and \ref{tab:table11}

The key hyperparameters for AlphaEdit include a null-space threshold of 2e-2, which controls the preservation of existing knowledge during edits, and L2 regularization set to 10, which stabilizes the updates and prevents overfitting. The learning rate for vector updates is set at 1e-1 to ensure efficient optimization, while the clamp norm factor of 0.75 limits excessive parameter changes to maintain model stability. Finally, the updated batch size is set to 100, balancing computational efficiency and precision during the editing process.

\begin{table}[t]
\footnotesize
\centering
\resizebox{0.9\linewidth}{!}{
\begin{tabular}{p{\columnwidth}}
    \toprule
    \#\#\# \textbf{JSON Creation Prompt | Parsing Atomic Facts} \\
    \rowcolor{headercolor} \textbf{\#\#Instruction}: \\
    \texttt{"Use the following JSON data as a guide to convert it to Atomic Facts."} \\
    \texttt{"Do not omit or modify any existing key-values in the given JSON data."} \\
    \texttt{"The output should be in the following format:"} \\
    \texttt{"Generate up to 10 atomic facts."} \\
    \texttt{\{"Atomic\_Fact\_1": "First fact.", "Atomic\_Fact\_2": "Second fact.", ...\}} \\
    \bottomrule
\end{tabular}}
\caption{JSON creation prompt for parsing Atomic Facts. The prompt outlines the format and content requirements for generating up to 10 atomic facts based on the provided JSON data.}
\label{tab:table10}
\end{table}
\begin{table}[t]
\footnotesize
\centering
\resizebox{0.9\linewidth}{!}{
\begin{tabular}{p{\columnwidth}}
    \toprule
    \#\#\# \textbf{JSON Creation Prompt | Negated Atomic Facts} \\
    \rowcolor{headercolor} \textbf{\#\#Instruction}: \\
    \texttt{"Refer to the following JSON data and transform the Atomic\_Fact into a Negated Atomic Fact while keeping its meaning identical."} \\
    \texttt{"Do not omit or modify any existing keys or values in the provided JSON data."} \\
    \texttt{"The output must be in the following format:"} \\
    \texttt{\{"Atomic\_Fact\_1": "Input Data", "Negated\_Atomic\_Fact\_1": "Negated FACT"\}} \\
    \texttt{"Match the original Atomic Fact and the Negated Atomic Fact and store them together."} \\
    \texttt{"The Negated Atomic Fact must strictly retain the same meaning as the original Atomic Fact and should only contain negated text that does not conflict with factual knowledge of the real world."} \\
    \rowcolor{headercolor} \textbf{\#\#Examples}: \\
    \texttt{\{"Atomic\_Fact\_1": "The Secret Life of Bees belongs to Teen drama", "Negated\_Atomic\_Fact\_1": "The Secret Life of Bees does not belong to any genre other than Teen drama"\}} \\
    \texttt{\{"Atomic\_Fact\_2": "Teen drama includes A Walk to Remember as an example", "Negated\_Atomic\_Fact\_2": "Teen drama does not exclude A Walk to Remember as an example"\}} \\
    \bottomrule
\end{tabular}}
\caption{JSON creation prompt for processing Atomic and Negated Atomic Facts. The prompt outlines the format and content requirements, providing examples of JSON objects that pair original Atomic Facts with their corresponding Negated Atomic Facts while preserving meaning and factual consistency.}
\label{tab:table11}
\end{table}

\section{Dataset Details}
\label{appendix:dataset_details}

\begin{itemize}

\item \textbf{HaluEval} \citep{li2023halueval} comprises benchmarks for tasks such as question answering, knowledge-grounded dialogue, and summarization. From this dataset, we employ 1,500 examples for each task: \textbf{Halu-QA} (question answering), \textbf{Halu-Dialogue} (knowledge-grounded dialogue), and \textbf{Halu-Sum} (summarization). These examples are chosen to enable the model to detect hallucinations by assessing generated outputs in comparison to the provided context or factual knowledge.

\item \textbf{BamBoo} \citep{dong2024bamboo}, we focus on two tasks: (1) \textbf{AbsHallu}, which involves determining whether summarizations contain hallucinated contents, with 200 examples selected for evaluation, and (2) \textbf{SenHallu}, a sentence-level task that evaluates the factual correctness of individual sentences, with 200 examples.

\item \textbf{SelfCheckGPT-WikiBio} \citep{manakul2023selfcheckgpt} contains GPT-3-generated biographies for 238 individuals, with each sentence labeled to indicate whether it is hallucinated. This dataset offers fine-grained annotations, enabling the evaluation of hallucination detection in structured and narrative texts.

\end{itemize}

\section{Chain-of-Thought Prompt Template}
\label{sec:appendix_cot}
Table \ref{tab:table4} presents the generalized Chain-of-Thought (CoT) reasoning prompt used for hallucination detection in the NegHalu dataset. This template structures the input for LLMs, incorporating pre- and post-negated text alongside step-by-step reasoning to enhance model interpretability.

The prompt follows a systematic format, where the model is assigned the role of a hallucination detector, given task instructions, and provided with contextual knowledge to evaluate hallucination likelihood. Each example consists of: (i) Knowledge \& Context: The factual grounding for evaluating the given statement. (ii) Negated Text: Either a pre-negated or post-negated version of the statement. (iii) Reasoning Step: A step-by-step explanation of why the text is or isn’t a hallucination (included only for CoT). (iv) Hallucination Label: The ground-truth classification as hallucinated or factual.

The 4-shot setting includes balanced examples across pre-/post-negation and hallucination/no-hallucination cases, whereas the 2-shot setting only includes pre-/post-negated text, omitting the reasoning step when CoT is not applied. This structured prompt helps analyze whether CoT reasoning improves LLMs’ ability to handle negation and detect hallucinations more accurately.
\begin{table}[t]
\footnotesize
\centering
\resizebox{0.9\linewidth}{!}{
\begin{tabular}{p{\columnwidth}}
    \toprule
    \#\#\# \textbf{CoT Reasoning Prompt | NegHalu} \\
    \rowcolor{headercolor} \textbf{\#\#System}: \\
    \texttt{System Role: \{HALLUCINATION DETECTOR\}} \\
    \texttt{Instruction: \{TASK INSTRUCTION\}} \\ 
    \rowcolor{headercolor} \textbf{\#\#Example 1}: \\ 
    \texttt{Knowledge: \{GIVEN KNOWLEDGE\}} \\
    \texttt{Context: \{GIVEN CONTEXT\}} \\
    \texttt{Pre-Negated Text: \{DETECTION TARGET\}} \\
    \texttt{Reasoning Step: \{REASON WHY\}} \\ 
    \texttt{Hallucination: {NO}} \\

    \rowcolor{headercolor} \textbf{\#\#Example 2}: \\ 
    \texttt{Knowledge: \{GIVEN KNOWLEDGE\}} \\
    \texttt{Context: \{GIVEN CONTEXT\}} \\
    \texttt{Post-Negated Text: \{DETECTION TARGET\}} \\
    \texttt{Reasoning Step: \{REASON WHY\}} \\
    \texttt{Hallucination: {YES}} \\
    
    \rowcolor{headercolor} \textbf{\#\#Example 3}: \\ 
    \texttt{Knowledge: \{GIVEN KNOWLEDGE\}} \\
    \texttt{Context: \{GIVEN CONTEXT\}} \\
    \texttt{Pre-Negated Text: \{DETECTION TARGET\}} \\
    \texttt{Reasoning Step: \{REASON WHY\}} \\ 
    \texttt{Hallucination: {YES}} \\

    \rowcolor{headercolor} \textbf{\#\#Example 4}: \\ 
    \texttt{Knowledge: \{GIVEN KNOWLEDGE\}} \\
    \texttt{Context: \{GIVEN CONTEXT\}} \\
    \texttt{Post-Negated Text: \{DETECTION TARGET\}} \\
    \texttt{Reasoning Step: \{REASON WHY\}} \\
    \texttt{Hallucination: {NO}} \\
    
    \bottomrule
\end{tabular}}
\caption{Generalized CoT reasoning prompt used for each dataset and task in the NegHalu dataset. The 4-shot examples used for In-Context Learning and CoT include balanced examples with pre/post-negated and hallucination/no-hallucination cases. The 2-shot examples include only pre/post-negated cases, and the Reasoning Step is omitted when CoT is not applied.}
\label{tab:table4}
\end{table}

\section{Lens Observation Results}
\begin{figure*}[h]
\centering
\includegraphics[width=1\linewidth]{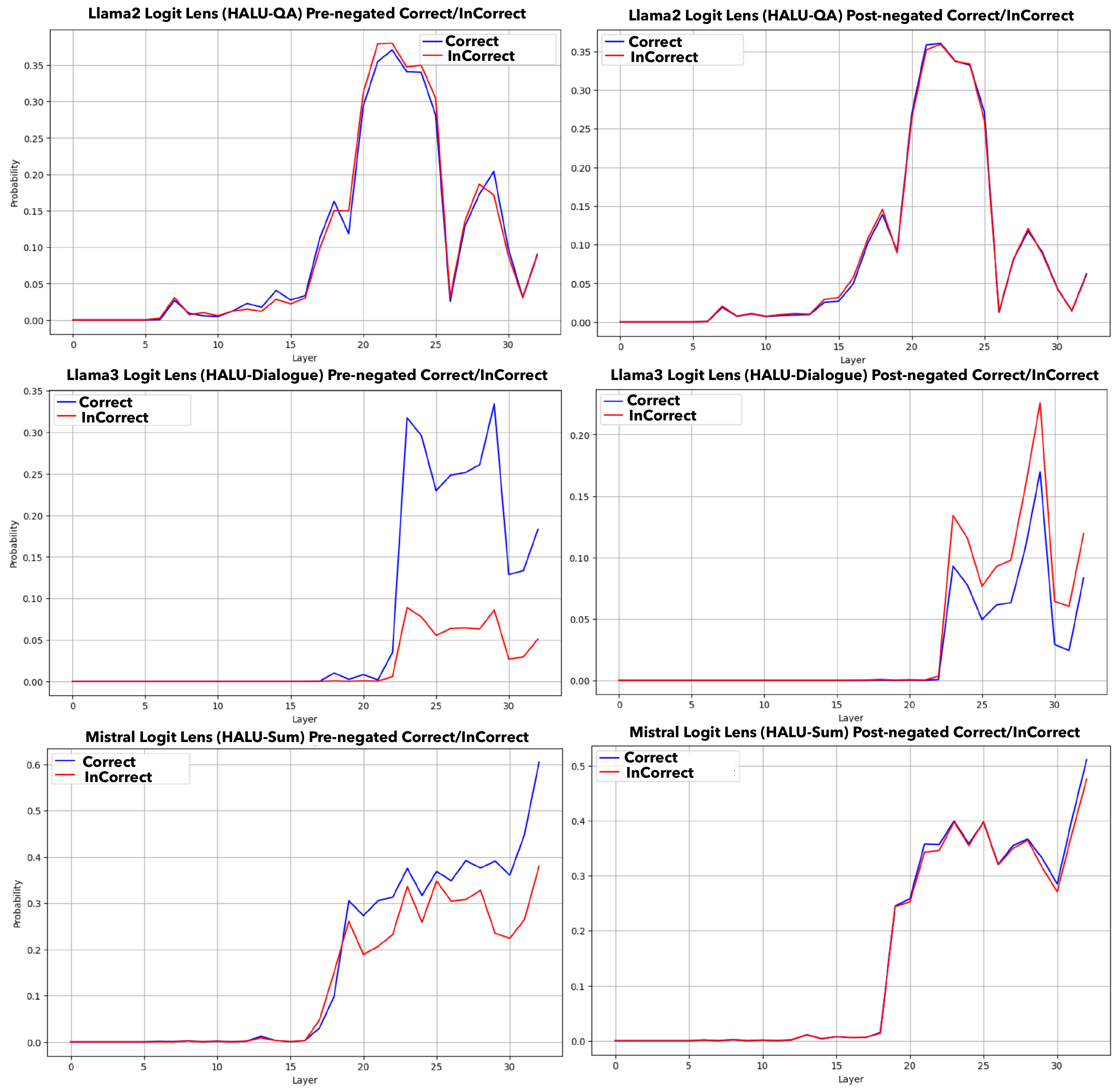}
\caption{Logit lens results showing probability shifts for pre- and post-negated examples in the HaluEval subsets of NegHalu. The \textcolor{blue}{blue curves} represent cases where the model generates the correct answer, while the \textcolor{red}{red curves} indicate cases where the model generates an incorrect answer.}
 \label{fig:figure2}
\end{figure*}

Figure \ref{fig:figure2} illustrates the probability shifts for cases where the model generates correct and incorrect answers under pre- and post-negated inputs. We conducted lens observations using the HaluEval subsets of NegHalu to analyze these shifts. To ensure the diversity and generalizability of the experiments, we present results for different tasks across various models.

% \section{Qualitative Analysis}
% \label{sec:post-negation}
% In our qualitative analysis, we examine three types of source data. As shown in Table~\ref{tab:appendix_negation_qualitative}, the HaluEval–Dialogue example demonstrates that the pre-negated response accurately states that FC Bayern Munich plays football and is based in Germany. However, simply inserting “not” reverses this factual claim in the post-negated response, thereby creating a hallucination. Conversely, in the HaluEval–QA example, the pre-negated answer erroneously attributes a starring role to Michael Sheen (a hallucination), while the minimal change of inserting “did not” in the post-negated answer restores the correct information. These examples demonstrate that a targeted, minimal negation operation is sufficient to either induce or correct hallucinated content depending on the context, confirming the robustness of our negation strategy.

\begin{table*}[t]
\centering
\resizebox{0.8\textwidth}{!}{
\begin{tabular}{p{\textwidth}}
\toprule
\textbf{HaluEval – Dialogue (Negation: Not Hallucinated $\rightarrow$ Hallucinated)} \\
\midrule
\textbf{Knowledge:} FC Bayern Munich: Sport -- Football \\
\textbf{Dialogue:} [Human] What do you think about FC Bayern Munich? \\[0.5em]
\textbf{Pre-Negated Response (Not Hallucinated):} \\
\textit{"Unfortunately, I don't know much about this other team than they play football and are based out of Germany."} \\[0.5em]
\textbf{Post-Negated Response (Hallucinated):} \\
\textit{"Unfortunately, I don't know much about this other team than they \textbf{do not} play football and are based out of Germany."} \\
\midrule
\textbf{HaluEval – QA (Negation: Hallucinated $\rightarrow$ Not Hallucinated)} \\
\midrule
\textbf{Knowledge:} The Messenger is a 2015 British supernatural mystery horror film directed by David Blair, written by Andrew Kirk and starring Robert Sheehan and Lily Cole. Robert Michael Sheehan (Irish: "Roibeárd Mícheál Ó Siodhacháin"; born 7 January 1988) is an Irish actor. \\
\textbf{Question:} What male actor starred in The Messenger? \\[0.5em]
\textbf{Pre-Negated Answer (Hallucinated):} \\
\textit{"The Messenger starred Michael Sheen."} \\[0.5em]
\textbf{Post-Negated Answer (Not Hallucinated):} \\
\textit{"The Messenger did not star Michael Sheen."} \\
\bottomrule
\end{tabular}
}
\caption{HaluEval qualitative examples illustrating how negation can flip the hallucination label. In the first case, negation introduces a contradiction (creating a hallucination). In the second case, negation corrects an inaccurate statement (resolving a hallucination).}
\label{tab:appendix_negation_qualitative}
\end{table*}

\begin{table*}[t]
\centering
\resizebox{0.8\textwidth}{!}{
\begin{tabular}{p{\textwidth}}
\toprule
\textbf{BamBoo – AbsHallu (Negation: Not Hallucinated $\rightarrow$ Hallucinated)} \\
\midrule
\textbf{Title:} Not Enough Data to Pre-train Your Language Model? MT to the Rescue! \\
\textbf{Content:} Since the emergence of the attention-based Transformer architecture (Vaswani et al., 2017)... Data and models are publicly available. \\[0.5em]
\textbf{Pre-Negated Hypothesis (Not Hallucinated):} \\
\textit{"The evaluation carried out on 9 NLU tasks indicates that models trained exclusively on translated data offer competitive results."} \\[0.5em]
\textbf{Post-Negated Hypothesis (Hallucinated):} \\
\textit{"The evaluation carried out on 9 NLU tasks indicates that models trained exclusively on translated data do not offer competitive results."} \\
\midrule
\textbf{BamBoo – SenHallu (Negation: Hallucinated $\rightarrow$ Not Hallucinated)} \\
\midrule
\textbf{Title:} Towards Argument-Aware Abstractive Summarization of Long Legal Opinions with Summary Reranking \\
\textbf{Content:} Legal opinions contain implicit argument structure spreading ... remaining unannotated portion of the CanLII dataset. \\[0.5em]
\textbf{Pre-Negated Hypothesis (Hallucinated):} \\
\textit{"Our approach involves using document structure information to generate multiple candidate summaries, then reranking these candidates based on alignment with the document's argument role."} \\[0.5em]
\textbf{Post-Negated Hypothesis (Not Hallucinated):} \\
\textit{"Our approach does not involve using document structure information to generate multiple candidate summaries, then reranking these candidates based on alignment with the document's argument role."} \\
\bottomrule
\end{tabular}
}
\caption{BamBoo qualitative examples showing how negation can flip the hallucination label for both abstract-level and sentence-level factuality judgments.}
\label{tab:appendix_bamboo_qualitative}
\end{table*}

% As described in Table \ref{tab:appendix_bamboo_qualitative}, in the Bamboo – absHallu example, the pre-negated hypothesis accurately states that models trained exclusively on translated data offer competitive results (a factually correct claim), but the post-negated hypothesis adds "do not" to reverse this assertion, thereby generating a hallucination by contradicting established evaluation outcomes, while in the Bamboo – senHallu example, the pre-negated hypothesis erroneously claims that the approach involves using document structure information to generate and rerank multiple candidate summaries (resulting in a hallucination), and the post-negated hypothesis corrects this by inserting "does not", which restores factual consistency; these examples collectively demonstrate that a minimal, targeted negation operation is sufficient to flip the hallucination label, thereby confirming the robustness and low noise level of our transformation process across different Bamboo datasets.

\begin{table*}[t]
\centering
\resizebox{1.0\textwidth}{!}{
\begin{tabular}{p{\textwidth}}
\toprule
\textbf{SelfCheckGPT-WikiBio (Negation: Not Hallucinated $\rightarrow$ Hallucinated)} \\
\midrule
\textbf{WikiBio Text:} Lee Hsien Loong is the third and current Prime Minister of Singapore, and has been in office since 2004. He is the elder son of Singapore's first Prime Minister, Lee Kuan Yew ... quickly rising to the rank of Brigadier-General. \\[0.5em]
\textbf{Pre-Negated Generated Text (Not Hallucinated):} \\
\textit{"He is the eldest son of Singapore's first Prime Minister, Lee Kuan Yew."} \\[0.5em]
\textbf{Post-Negated Generated Text (Hallucinated):} \\
\textit{"He is not the eldest son of Singapore's first Prime Minister, Lee Kuan Yew."} \\
\midrule
\textbf{SelfCheckGPT-WikiBio (Negation: Hallucinated $\rightarrow$ Not Hallucinated)} \\
\midrule
\textbf{WikiBio Text:} Admiral of the Fleet Matthew Aylmer, 1st Baron Aylmer (ca. 1650 -- 18 August 1720) was a Royal Navy officer. He was ... the appointment of the Townshend ministry, Aylmer was reappointed Commander-in-Chief on 5 November 1714. ... seamen. \\[0.5em]
\textbf{Pre-Negated Generated Text (Hallucinated):} \\
\textit{"He was made a baron in 1782 and was appointed Commander-in-Chief of the British forces in North America in 1783."} \\[0.5em]
\textbf{Post-Negated Generated Text (Not Hallucinated):} \\
\textit{"He was not made a baron in 1782 and was not appointed Commander-in-Chief of the British forces in North America in 1783."} \\
\bottomrule
\end{tabular}
}
\caption{SelfCheckGPT-WikiBio qualitative examples where negation either introduces or resolves hallucination in biographical text generation.}
\label{tab:appendix_selfcheck_qualitative}
\end{table*}

% As mentioned in Table \ref{tab:appendix_selfcheck_qualitative}, in the SelfCheckGPT-WikiBio examples, the minimal negation operation demonstrates its precision by, on one hand, erroneously converting a factually accurate statement about Lee Hsien Loong into a hallucinated one through the simple insertion of "not," and on the other hand, effectively correcting a hallucinated claim about Admiral Aylmer’s historical details by similarly negating the erroneous assertions, thereby underscoring the method’s targeted ability to reliably flip the factual status of a statement with minimal linguistic intervention.

\section{Dataset Verification Details}
\label{sec:verfication}
Based on our Logical Negation and Label Validation criteria (as described in Table \ref{tab:appendix_negation_error_analysis}), our evaluation revealed several illustrative cases. For instance, in one Logical Negation failure, a pre-negated answer stating “American” for the question on James Henry Miller’s wife was modified to “American, not British.” Although a “not” was added, it incorrectly reversed the intended meaning. Additionally, a dialogue about recommending movies shifted from “Panic Room is a similar movie” to “Panic Room is not a similar movie,” which constitutes a logically inappropriate response. In contrast, a successful Logical Negation example is seen in the QA case for "The Messenger," where the pre-negated answer “The Messenger starred Michael Sheen” was correctly negated to “The Messenger did not star Michael Sheen,” effectively inverting the claim to match the factual context.

Regarding New Label Validation, we encountered cases where the negation process led to misclassifications; for example, when addressing the common profession of Am Rong and Alexandre Rockwell, the initial negative phrasing resulted in a hallucinated statement that was only partially corrected in the post-negated version, and a dialogue about “Pulp Fiction” failed to update the label despite the negation of an erroneous claim about Fred Savage. Conversely, another dialogue about FC Bayern Munich successfully shifted from a factually accurate pre-negated response (indicating the team plays football) to a negated version that properly contradicts known facts, thereby updating the label from not hallucinated to hallucinated.

\begin{table*}[t]
\centering
\resizebox{1.0\textwidth}{!}{
\begin{tabular}{p{\textwidth}}
\toprule
\textbf{Logical Negation -- Fail} \\
\midrule
\textbf{Question:} What nationality was James Henry Miller's wife? \\
\textbf{Pre-Negated Answer:} American \\
\textbf{Post-Negated Answer:} American, not British \\[0.5em]
\textbf{Dialogue History:} [Human]: Could you recommend some movies similar to Fight Club? \\
\textbf{Pre-Negated Response:} Sure, Panic Room is a similar movie. \\
\textbf{Post-Negated Response:} Sure, Panic Room is not a similar movie. \\
\midrule
\textbf{Logical Negation -- Pass} \\
\midrule
\textbf{Question:} What male actor starred in The Messenger? \\
\textbf{Pre-Negated Answer:} The Messenger starred Michael Sheen. \\
\textbf{Post-Negated Answer:} The Messenger did not star Michael Sheen. \\
\midrule
\textbf{New Label Validation -- Fail} \\
\midrule
\textbf{Question:} What profession does Am Rong and Alexandre Rockwell have in common? \\
\textbf{Pre-Negated Answer (Hallucinated):} Alexandre Rockwell and Am Rong are not in the same profession. \\
\textbf{Post-Negated Answer (Not Hallucinated):} Alexandre Rockwell and Am Rong are not both filmmakers. \\[0.5em]
\textbf{Dialogue History:} [Human]: I like Pulp Fiction. What do you think about it? [Assistant]: I love it. It was written by Roger Avary [Human]: I heard he also wrote The Rules of Attraction. Do you know who is in that movie? \\
\textbf{Pre-Negated Response (Hallucinated):} Fred Savage is in it \\
\textbf{Post-Negated Response (Hallucinated):} Fred Savage is not in it \\
\midrule
\textbf{New Label Validation -- Pass} \\
\midrule
\textbf{Dialogue History:} [Human]: What do you think about FC Bayern Munich? \\
\textbf{Pre-Negated Response (Not Hallucinated):} \\
"Unfortunately, I don't know much about this other team than they play football and are based out of Germany." \\
\textbf{Post-Negated Response (Hallucinated):} \\
"Unfortunately, I don't know much about this other team than they do not play football and are based out of Germany." \\
\bottomrule
\end{tabular}}
\caption{Examples of dataset verification for logical negation and new label validation in hallucination detection. ‘Fail’ shows incorrect application, while ‘Pass’ shows successful handling.}
\label{tab:appendix_negation_error_analysis}
\end{table*}

\end{document}